\documentclass[conference]{IEEEtran}
\usepackage{times}
\usepackage[acronym]{glossaries}

\newacronym{bc}{BC}{Behavior Cloning}
\newacronym{llms}{LLMs}{Large Language Models}
\newacronym{vr}{VR}{Virtual Reality}
\newacronym{mpc}{MPC}{Model Predictive Control}

\usepackage[numbers]{natbib}
\usepackage{multicol}
\usepackage[bookmarks=true]{hyperref}
\usepackage{amsmath}
\usepackage{comment}
\usepackage{graphicx}
\usepackage{color,xcolor,soul}
\usepackage{multirow}
\usepackage{multicol}
\usepackage{makecell}
\usepackage{tabularx}
\usepackage{algorithmic}
\usepackage[ruled,vlined,linesnumbered]{algorithm2e}
\usepackage{color-edits}
\usepackage[bookmarks=true]{hyperref}
\usepackage{booktabs}
\usepackage[font=small]{caption}
\usepackage{amsfonts}
\usepackage{threeparttable}

\addauthor{bpg}{purple}
\addauthor{lu}{red}
\addauthor{pang}{green}
\addauthor{tong}{orange}
\addauthor{russt}{CornflowerBlue}

\DeclareMathOperator*{\argmin}{arg\,min}

\newcommand\extrafootertext[1]{%
    \bgroup
    \renewcommand\thefootnote{\fnsymbol{footnote}}%
    \renewcommand\thempfootnote{\fnsymbol{mpfootnote}}%
    \footnotetext[0]{#1}%
    \egroup
}

\begin{document}

\title{Physics-Driven Data Generation for Contact-Rich Manipulation via Trajectory Optimization
}

\author{
Lujie Yang$^{1, 2}$, H.J. Terry Suh$^{* 1}$, Tong Zhao$^{* 2}$, Bernhard Paus Gr\ae sdal$^1$, Tarik Kelestemur$^2$,
\\Jiuguang Wang$^2$, Tao Pang$^2$, and Russ Tedrake$^1$
\thanks{*Equal contributions. $^{1}$Computer Science and Artificial Intelligence Laboratory (CSAIL), Massachusetts Institute of Technology. $^{2}$Robotics and AI Institute. Correspondence to {\tt\small lujie@mit.edu}.} \\
\authorblockA{$^1$Massachusetts Institute of Technology \quad $^{2}$Robotics and AI Institute }
}

\vspace{1.5cm}
\makeatletter
\let\@oldmaketitle\@maketitle
\renewcommand{\@maketitle}{\@oldmaketitle
  \captionsetup{type=figure}
  \includegraphics[width=1.0\linewidth]
    {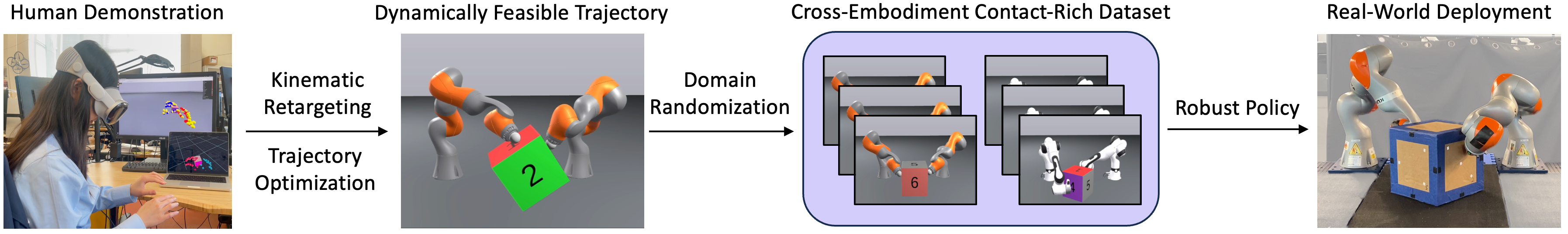}
    \captionof{figure}{\small \textbf{Physics-driven data generation overview.} Leveraging trajectory optimization, our framework automatically generates thousands of dynamically feasible contact-rich trajectories across a range of embodiments and physical parameters from only 24 human demonstrations. The policy trained with imitation learning from the generated dataset is more robust and performant.
    }
    \label{fig:banner}
    }%
\makeatother

\maketitle


\begin{abstract}
We present a low-cost data generation pipeline that integrates physics-based simulation, human demonstrations, and model-based planning to efficiently generate large-scale, high-quality datasets for contact-rich robotic manipulation tasks.
Starting with a small number of embodiment-flexible human demonstrations collected in a virtual reality simulation environment, the pipeline refines these demonstrations using optimization-based kinematic retargeting and trajectory optimization to adapt them across various robot embodiments and physical parameters.
This process yields a diverse, physically consistent, contact-rich dataset that enables cross-embodiment data transfer, and offers the potential to reuse legacy datasets collected under different hardware configurations or physical parameters.
We validate the pipeline’s effectiveness by training diffusion policies from the generated datasets for challenging long-horizon contact-rich manipulation tasks across multiple robot embodiments, including a floating Allegro hand and bimanual robot arms. The trained policies are deployed zero-shot on hardware for bimanual iiwa arms, achieving high success rates with minimal human input. Project website: \href{https://lujieyang.github.io/physicsgen/}{https://lujieyang.github.io/physicsgen/}.
\end{abstract}

\extrafootertext{* Equal contribution. Correspondence to \texttt{$<$lujie@mit.edu$>$}.}

\section{Introduction}

The emergence of foundation models has transformed fields such as natural language processing and computer vision, where models trained on massive, internet-scale datasets demonstrate remarkable generalization across diverse reasoning tasks \cite{achiam2023gpt, touvron2023llama, anil2023palm}. Motivated by this success, the robotics community is currently pursuing foundation models for generalist robot policies capable of flexible and robust decision-making across a wide range of tasks \cite{kim2024openvla, o2023open, team2024octo}, leading to significant industrial investments in large-scale robot learning \cite{reuters_figure_funding_2024}.
However, the pursuit for generalist robot policies remains constrained by the limited availability of high-quality datasets, especially for contact-rich robotic manipulation. Existing datasets \cite{o2023open, khazatsky2024droid, walke2023bridgedata, dasari2019robonet} are orders of magnitude smaller than those used to train foundation models in other domains, such as \acrfull*{llms}. The scarcity of diverse, high-fidelity manipulation data limits policy generalization across different embodiments, task contexts, and physical conditions. 


To address data scarcity, robot learning researchers often rely on a spectrum of data sources varying in cost, quality, and transferability. The most informative data typically consists of high-quality demonstrations specific to the task, environment, and embodiment \cite{o2023open, khazatsky2024droid}, but such data is costly and time-consuming to collect, as it requires human teleoperation with specialized hardware. At the opposite end of the spectrum, there is a wealth of lower-quality data in the form of internet videos showing humans and robots performing manipulation tasks \cite{damen2018scaling, radosavovic2023real, nair2022r3m, karamcheti2023language}. However, the significant embodiment gap and limited action labeling make this data difficult to transfer effectively to robot policies. Simulation data offers a middle ground, providing the potential to generate large, diverse, and high-quality datasets at relatively low cost \cite{xiang2020sapien, brockman2016openai, james2020rlbench}. In practice, effective policy learning can be achieved by co-training on a mixture of data from different points along this spectrum, reducing data collection costs while improving generalization \cite{wang2024scaling, wei2025empirical}. \looseness=-1



A key insight in this work is that human demonstrations and model-based planners complement each other in critical ways for generating high-quality robot data. 
Human demonstrations, though costly to collect, offer valuable global information for solving complex tasks.
However, collecting real-world, contact-rich manipulation data through teleoperation is challenging due to the need for precise multi-contact interactions, which are difficult to achieve in practice due to hardware latency, embodiment mismatches between the human and robot, and the fine-grained control required \cite{chi2024universal}. 
In contrast, trajectory optimization has demonstrated success in generating locally-optimal trajectories for contact-rich tasks \cite{mordatch2012discovery, posa2014direct, howell2022calipso}, but often relies on global guidance in the form of good initial guesses. \looseness=-1
In this work, we propose a data generation framework that leverages the strengths of both approaches: human demonstrations can provide global guidance, while trajectory optimization can locally refine these demonstrations to ensure dynamic feasibility. 
Starting with a small number of human demonstrations collected in a virtual reality (VR) environment, our method uses model-based trajectory optimization to generate large datasets of dynamically feasible, contact-rich trajectories in simulation. The demonstrations guide the planner through complex search spaces, while the planner ensures physical consistency and robustness across varying physical parameters and robot embodiments.
Our pipeline, visualized in Fig. \ref{fig:banner}, enables efficient cross-embodiment data transfer, where demonstrations collected with one robot configuration can be adapted to another, and supports domain randomization for improved generalization and robustness.
Additionally, it provides the potential to revive and adapt legacy datasets collected with different hardware or configurations, making old datasets valuable for new robot systems. \looseness=-1

Our key contributions include:
\begin{enumerate}
    \item We present an intuitive, embodiment-flexible demonstration interface based on virtual reality and physics simulation, enabling fast data collection for dexterous contact-rich manipulation.  
    \item  We propose a scalable framework that leverages trajectory optimization to transform a small number of human demonstrations into large-scale, physically consistent datasets, enabling generalization across embodiments, initial conditions, and physical parameters.
    \item We validate our approach by training policies on the generated dataset for challenging contact-rich manipulation tasks across multiple robot platforms, including bimanual robot arms and a floating base Allegro hand.
    \item We achieve high success rates in zero-shot hardware deployment on bimanual iiwa arms, highlighting the utility of augmented datasets in real-world scenarios.
    
\end{enumerate}

\section{Related Works}
In this section, we review the most relevant approaches for generating diverse robot data for contact-rich tasks.
We categorize the methods into data collection, data augmentation, model-based planning, demonstration-guided reinforcement learning and cross-embodiment transfer.

\subsection{Data Collection for Imitation Learning}
Behavior Cloning \cite{pomerleau1988alvinn}, which trains robot policies to mimic expert behavior, has shown impressive empirical results in a wide range of dexterous manipulation tasks \cite{chi2023diffusion}. Collecting high-quality robot data has been an essential component of imitation learning (IL). Many such methods rely on human experts teleoperating a robot to accomplish specific tasks. Researchers have adopted interfaces such as 3D spacemouse \cite{chi2023diffusion, zhu2023viola}, 
and puppeteering platforms \cite{fu2024mobile, zhao2024aloha} for end-effector \cite{o2023open} and whole-body control \cite{fu2024humanplus, he2024omnih2o}.


Virtual and augmented reality (VR/AR) interfaces have recently gained traction as effective alternatives for robot data collection \cite{ebert2021bridge, zhang2018deep}, reducing cognitive load, physical strain, and user frustration compared to traditional techniques like kinesthetic teaching or 3D mouse control \cite{smith2024augmented}. These technologies offer a more intuitive data collection paradigm for complex tasks, especially in dexterous manipulation. AR2-D2 \cite{duan2023ar2} enables data collection without a physical robot by projecting a virtual robot into the physical workspace, but lacks real-time feedback necessary for precise control. 
DART~\cite{park2024dexhub} supports data collection entirely in simulation, visualized through a VR headset, but faces challenges bridging the sim-to-real gap for physical robot deployment.  
ARCap~\cite{chen2024arcap} integrates real-time AR feedback, but requires specialized hardware, including an RGBD camera, motion capture gloves, and VR controllers, in addition to the AR headset. ARMADA~\cite{nechyporenko2024armada} enables real-world manipulation data collection with bare hands through real-time virtual robot feedback, achieving high success rates when replayed on physical hardware. In contrast to these existing systems, our work focuses on scalable data generation from a small number of human demonstrations by leveraging trajectory optimization, facilitating generalization across different robot embodiments, initial conditions, and physical parameters.\looseness=-1

\subsection{Data Augmentation}
Despite many research efforts, collecting large datasets remains time-consuming and costly, requiring a large amount of human effort and resources. To address these challenges, significant effort has been devoted to automating the data generation process through data augmentation techniques. Existing approaches have leveraged state-of-the-art generative models for visual \cite{zhang2024diffusion, tian2024view, chen2024rovi} and semantic \cite{mandi2022cacti, chen2023genaug, yu2023scaling} augmentations. MimicGen \cite{mandlekar2023mimicgen} and its bimanual extension DexMimicGen \cite{ jiang2024dexmimicgen} automatically synthesize large-scale datasets from a small number of human demonstrations. These works decompose long-horizon tasks into object-centric subtasks and replay transformed demonstrations open loop in simulation. SkillMimicGen \cite{garrett2024skillmimicgen} extends this paradigm by segmenting tasks into motion and skill components, augmenting local manipulation skills with MimicGen-style replay and using motion planning to connect these skill segments.  RoboCasa \cite{nasiriany2024robocasa} leverages generative models to create diverse kitchen scenes with abundant 3D assets and utilizes MimicGen for automated trajectory generation. While these approaches have shown success in automating data generation, they primarily rely on kinematic replay of demonstrations, which is often inadequate for contact-rich manipulation tasks. Our work can be viewed as an important extension to MimicGen line of works to support dynamically feasible contact-rich data generation, which requires fine-grained control of the robot and continuous reasoning about making and breaking contacts with the environment.

\subsection{Trajectory Optimization for Contact-Rich Tasks}


Planning and control through contact remains a significant challenge for both learning-based and model-based methods due to the explosion of contact modes and the nonsmooth nature of contact dynamics. To tackle these challenges, researchers have explored various trajectory optimization formulations for multi-contact interactions.


\noindent \textbf{Contact-Implicit Trajectory Optimization} Existing works based on contact-implicit trajectory optimization (CITO) \cite{posa2014direct, mordatch2012discovery} have sought to formulate the combinatorial problem into a smooth optimization problem by using complementarity constraints. CITO has been applied in various domains, including planar manipulation \cite{onol2019contact, moura2022non}, dynamic pushing \cite{sleiman2019contact}, and locomotion tasks \cite{tassa2012synthesis, neunert2016efficient, winkler2018gait}. Recent efforts have extended CITO for real-time applications as model predictive control (MPC) \cite{wang2022contact, kurtz2023inverse}, with successful hardware deployment on quadrupeds using tailored solvers \cite{neunert2018whole, le2024fast}.  
Aydinoglu et al. \cite{aydinoglu2024consensus} parallelize the solution of linear complementarity problems using alternating direction method of multipliers (ADMM) and validate the method on hardware for multi-contact manipulation tasks.
While CITO shows promising scalability for handling contact modes, it suffers from poor global exploration and relies on good initial guesses \cite{ctr}.
A new line of work tries to address these issues with efficient global optimization \cite{graesdal2024towards}, but does not yet scale to the tasks we consider here.

\noindent \textbf{Sampling-Based Planning} Sampling-based methods have also shown great promise for solving trajectory optimization for contact-rich tasks. H{\"a}m{\"a}l{\"a}inen et al. \cite{hamalainen2015online} employ sampling-based belief propagation for humanoid balancing, juggling and locomotion. Carius et al. 
\cite{carius2022constrained} extend the path integral formulation to handle state-input constraints and validate the approach on quadruped stabilization on hardware.
More recently, Pezzato et al. \cite{pezzato2023sampling} applied sampling-based predictive control (SPC) for simpler contact tasks like pushing, while Howell et al. \cite{howell2022predictive} and Li et al. \cite{li2024drop} extended SPC to more complex, contact-rich tasks such as in-hand cube reorientation. Pang et al. \cite{pang2023global} use smoothed contact dynamics with global sampling to generate contact-rich plans in under a minute, with performance comparable to reinforcement learning. Cheng et al. introduce HiDex \cite{cheng2023enhancing}, a hierarchical planner that combines Monte-Carlo Tree Search with integrated contact projection, achieving rapid planning for dexterous manipulation tasks. 
Interestingly, directly applying sampling-based planners for contact-rich data generation in behavior cloning can be problematic, as the high entropy of the generated trajectories often degrades downstream policy performance \cite{belkhale2024data, zhu2024should}.

In this work, we leverage low-entropy human demonstrations to guide the global planning for multi-contact interactions and utilize trajectory optimization to locally refine the trajectories for specific physical parameters and robot embodiments. From a small number of demonstrations, the model-based planner can efficiently generate abundant, high-quality, contact-rich data for training robust robot policies.

\subsection{Demonstration-Guided Reinforcement Learning}
While IL often demands a large number of expert demonstrations to achieve robust and high-performing policies, reinforcement learning (RL) aims to solve tasks autonomously through reward-driven exploration. However, pure RL can suffer from inefficient exploration and the need for extensive reward shaping, especially in complex manipulation tasks \cite{chen2022system, qi2023hand}. To address these challenges, researchers have explored using demonstrations to guide RL, improving both sample efficiency and exploration quality.

Demonstrations have been integrated into RL pipelines in various ways, including adding them directly to the replay buffer \cite{vecerik2017leveraging, nair2018overcoming}, using behavior cloning for policy pretraining \cite{rajeswaran2017learning, hu2023imitation, hansen2022modem}, and augmenting task rewards with information extracted from demonstrations \cite{zhu2018reinforcement, peng2021amp, peng2018deepmimic}. Sleiman et al. \cite{sleiman2024guided} guide RL with demonstrations generated from a model-based trajectory optimizer for multi-contact loco-manipulation tasks, and validate their method on hardware with a quadrupedal mobile manipulator. While these approaches search over the parameters of a neural network policy and potentially optimize a more global objective, we leverage trajectory optimization as a complementary tool to \emph{locally} refine and expand demonstration trajectories. This enables the efficient generation of contact-rich data while avoiding the computational overhead, approximation errors, and unnecessary exploration associated with RL's high-dimensional search space.

\subsection{Cross-Embodiment Generalization}
Reusing datasets and policies across different embodiments unlocks the potential for large-scale robot learning. One line of work learns latent plans from videos of humans interacting with the environment and transfers this knowledge for robotic manipulation \cite{li2024okami, wang2023mimicplay}. Another approach involves portable data collection tools, such as hand-held grippers \cite{chi2024universal, seo2024legato}, for in-the-wild human demonstrations. While these methods enable policy deployment on multiple robot platforms, they are often constrained to robots with the same end-effector used during data collection, limiting generalization across platforms. On the other hand, to leverage large-scale datasets, recent works pull data from a heterogeneous set of robots ranging from navigation to manipulation, and train a robotic foundation model capable of accomplishing a diverse range of tasks \cite{yang2024pushing, doshiscaling}. Our proposed framework enables reusing the same set of easy-to-collect demonstrations for multiple robots, avoiding the need to collect embodiment-specific data for contact-rich tasks.

\section{Data Collection}

We present a Virtual Reality (VR)-based data collection pipeline designed for intuitive and efficient collection of human demonstrations across multiple robot embodiments. 
The pipeline emphasizes simplicity and cross-embodiment generalization while minimizing the reliance on physical robot hardware. 
While we consider the data collection pipeline to be one of our contributions, we emphasize that the simulation-based large-scale data generation method presented in the next section is independent of this particular data collection approach. \looseness=-1

\begin{figure}
\centering
	\includegraphics[width=0.42\textwidth]{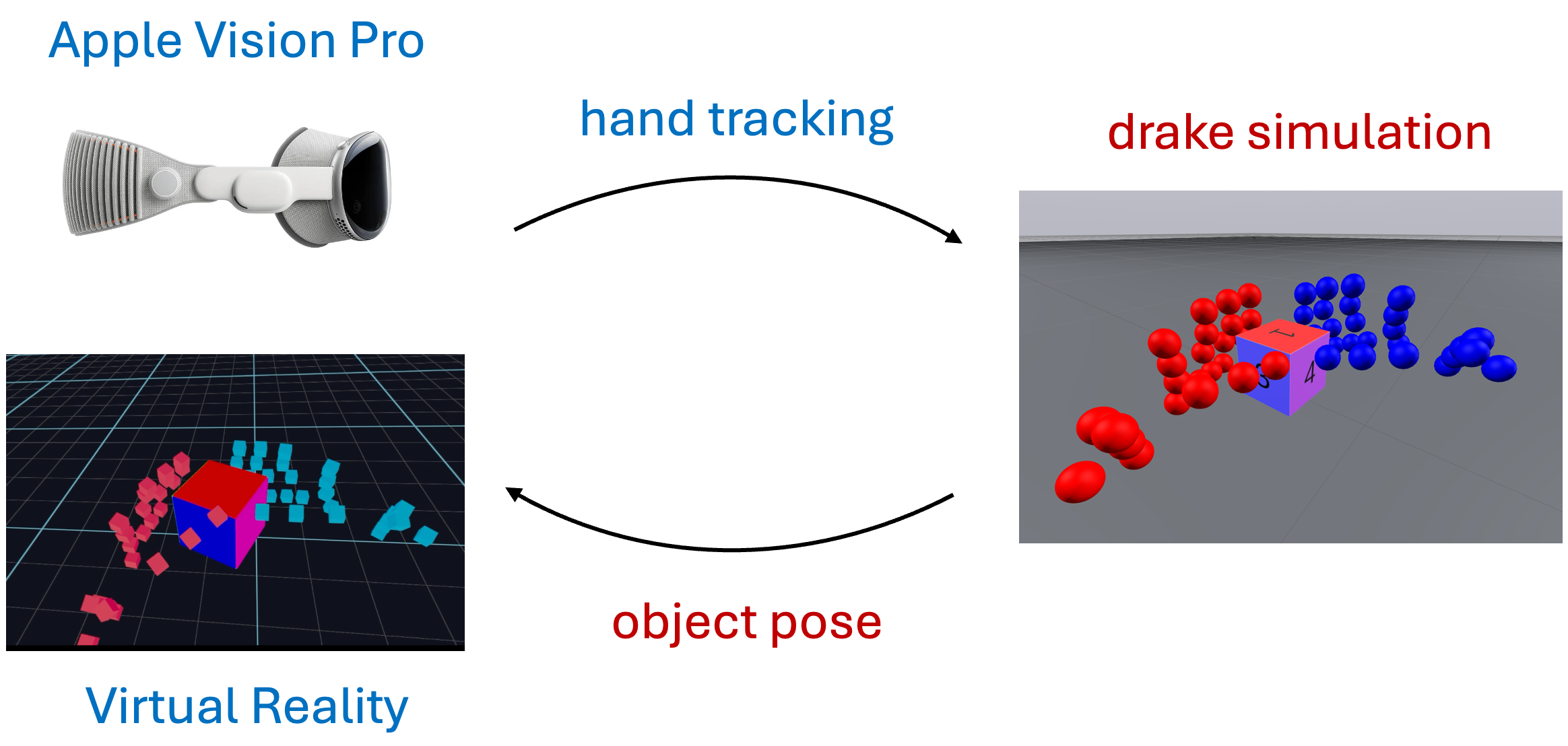}
	\caption{VR-based human-hand demonstration framework.}
	\label{fig:data_collection}
\end{figure}
\begin{figure*}[t]
\centering
	\includegraphics[width=1.0\textwidth]{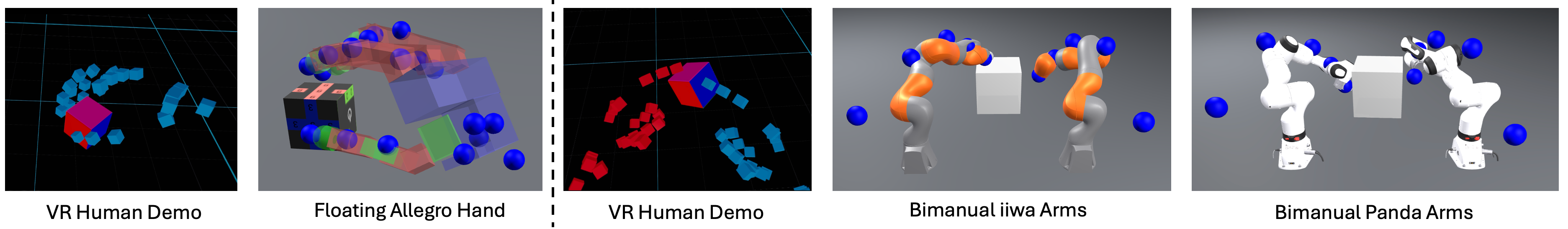}
	\caption{Human hand demo in VR and kinematic retargeting for different embodiments. The blue spheres illustrate the demo hand landmarks scaled to the specific system.} 
	\label{fig:kinematic_retargeting}
	\vspace*{-0.2cm}
\end{figure*}
Our data collection pipeline (Fig. \ref{fig:data_collection}) is a human-hand demonstration interface in VR. We use an Apple Vision Pro to track the poses of the human demonstrator's hands and stream the poses to the Drake physics simulator \cite{drake}, which simulates the contact interaction between the object and the hands. The updated object pose is then sent back to Apple Vision Pro for real-time visualization in VR using Vuer~\cite{vuer}.

Our demonstration interface is fast and cost-effective. Since the system operates entirely in simulation, it removes the dependency on robot hardware, significantly reducing the cost and complexity of data collection. 
In practice, it takes approximately 7 minutes to collect 24 long-horizon demos for each considered system.
The setup is also intuitive to use, as the human demonstrator does not have to mentally close the embodiment gap between the human body and the specific robot.\looseness=-1

We demonstrate our pipeline on two different classes of robot embodiments: a dexterous hand and a bimanual manipulation setup.

\underline{\textbf{Floating Allegro Hand}} For the dexterous hand, we consider a 22-DOF free-floating Allegro hand manipulating a cube on a table as shown in Fig. \ref{fig:kinematic_retargeting}. Since the Allegro hand only has four fingers, we restrict the VR-based demonstrations to using four fingers on the right hand to interact with the object in simulation.

\underline{\textbf{Bimanual Robot Arms}} For the bimanual manipulation setup, we consider two different fixed-base bimanual manipulators: a pair of 7-DOF Kuka LBR iiwa arms, and a pair of Franka Emika Panda arms. Each pair of arms collaboratively manipulates a big box (Fig. \ref{fig:kinematic_retargeting}). During the VR demonstrations, the human demonstrator uses both index fingers to manipulate a small cube in VR and constrains their wrist movement to mimic the fixed base. During kinematic motion retargeting (detailed in Sec. \ref{subsec:kin_retarget}), the small cube and fingers are scaled to match the size of the larger box and the robot manipulators. 

This design facilitates two forms of cross-embodiment generalization. First, it utilizes easy-to-collect human finger demonstrations to guide planning for harder and higher-dimensional tasks, such as the dual-arm manipulators. Second, it supports the reuse of the same set of demonstrations across multiple robot platforms, as both the iiwa and Panda arms can leverage the same data to accomplish the manipulation task, eliminating the need for embodiment-specific demonstrations.


\section{Automated Data Generation}
In this section, we present our method for automatically generating large quantities of physically feasible trajectories for contact-rich manipulation tasks across a range of objects, initial conditions, and embodiments from only a handful of demonstrations.
The presented method also offers the potential to adapt legacy datasets collected using outdated configurations to new robot settings, reducing the cost of collecting large amounts of data on the new robot setups from scratch.

Our method starts out by retargeting kinematic motions from the original embodiment-flexible human demonstrations collected in VR to the specific robot embodiment in simulation, producing kinematically feasible trajectories. These trajectories are then refined and augmented through the use of local trajectory optimization to obtain dynamically feasible trajectories for a range of physical parameters. The following subsections provide a detailed breakdown of each step in the pipeline. \looseness=-1

\subsection{Kinematic Motion Retargeting}
\label{subsec:kin_retarget}
Given a sequence of demonstrations $x^{\text{demo}}_{0:T}$ with horizon $T$, we aim to find the robot configurations $q^{\text{retarget}}_{0:T}$ that match the positioning of the demonstrator while avoiding penetration and obeying joint limits. At each time step, we solve the following nonconvex program:
\begin{subequations}\label{eq:retargeting_ik}
    \begin{align}
        {q_t^{\text{retarget}}}^\star = \argmin_{q_t^{\text{retarget}}} \quad &\sum_{i=0}^N w_i \|\psi_i(q_t^{\text{retarget}}) - \Tilde \psi_i(x^{\text{demo}}_t)\|^2 \label{eq:match_cost}\\
        \text{s.t. } \; &\phi_j(q_t^{\text{retarget}}) \ge 0, \; \forall j \label{eq:nonpenetration_constr}\\ 
        &q_{\text{min}} \le q_t^{\text{retarget}} \le q_{\text{max}},
    \end{align}
\end{subequations}
where $w_i>0$ are weight parameters, and $\psi_i$ and $\Tilde \psi_i$ represent the $i$-th mappings from the robot configuration and demonstrator state to corresponding points on the embodiments. The corresponding points of interest for each robot/demonstrator pair are manually defined. For example, on the bimanual robot arm system, $\psi_0$ is the forward kinematics from the robot joint angles to the left robot arm's end effector position, while $\Tilde \psi_0$ is a map from the hand pose to the fingertip of the left index finger. We find the resulting plans generated by trajectory optimization relatively robust to the correspondence and weight parameter selection. $\phi_j$ denotes the signed distance function between the $j$-th collision pair and \eqref{eq:nonpenetration_constr} enforces nonpenetration constraints. $q_{\text{min}}$ and $q_{\text{max}}$ are the lower and upper bounds on the joint angles. Notice that $q^{\text{retarget}}$ and $x^{\text{demo}}$ can have different dimensions as long as both $\psi_i$ and $\Tilde \psi_i$ map them to vectors in the same space (e.g., Apple Vision Pro captures 5 landmarks on the index finger while each robot arm has 7 DOF in the bimanual robot arm system). We solve \eqref{eq:retargeting_ik} using a Sequential Quadratic Programming (SQP)-style algorithm: during each iteration, the nonpenetration constraint \eqref{eq:nonpenetration_constr} is linearized and the matching objective \eqref{eq:match_cost} is quadratically approximated around the solution to the previous iteration. We warmstart the solution of the nonlinear program at time $t$ with the optimal solution from the previous timestep ${q_{t-1}^{\text{retarget}}}^\star$ to encourage faster convergence and temporal consistency.

\subsection{Demonstration-Guided Trajectory Optimization}
The kinematically consistent robot trajectories ${q_{0:T}^{\text{retarget}}}^\star$ are generally not dynamically feasible due to the embodiment gap and differences in physical parameters. However, they can provide good guidance on generating dynamically feasible trajectories with complex multi-contact interactions.
In particular, human demonstrations provide global information about when and where to make contact with the object, which model-based planning can then locally refine. We define the retargeted system state $x_t^{\text{retarget}}$ to incorporate both the object state $x_t^{\text{object}}$, which is a subset of $x_t^{\text{demo}}$, and the robot state as a function of ${q_{t}^{\text{retarget}}}^\star$. The trajectory $x_{0:T}^{\text{retarget}}$ is then locally refined by solving the following nonconvex optimization program: 
\begin{subequations}\label{eq:predictive_control}
    \begin{align}
        x_t^\star, u_t^\star = \argmin_{x_t, u_t} \quad &\|x_T - {x_T^{\text{retarget}}}\|_{Q_T}^2 \nonumber + \\
         \sum_{t=0}^{T-1} &(\|x_t - {x_t^{\text{retarget}}}\|_{Q_t}^2 + \|u_t\|_{R_t}^2) \label{eq:predictive_control_cost}\\
        \text{s.t. } \; &x_{t+1} = f(x_t, u_t) \label{eq:pc_dynamics}\\
        &\phi_j(x_t) \ge 0, \; \forall j \\ 
        &x_{\text{min}} \le x_t \le x_{\text{max}} \\
        &u_{\text{min}} \le u_t \le u_{\text{max}}.
    \end{align}
\end{subequations}
Here, $u$ is the control input, $f$ is obtained by time-stepping the dynamics engine, $x_{\text{min}} / x_{\text{max}}$ ($u_{\text{min}} / u_{\text{max}}$) are the lower and upper bounds on the state (input), $Q_t, R_t$ are the cost matrices for the state and input, respectively, and $Q_T$ is the cost matrix for the terminal state. To encourage precise tracking of the object trajectory, we assign higher weights to the entries of  $Q_t$ which correspond to $x_t^{\text{object}}$. The detailed parameters can be found in Appendix \ref{sec:appendix_cem}.

In general, model-based planners can struggle to discover high-quality long-horizon contact-rich trajectories without demonstrations. 
CITO requires good initial guesses and can easily get stuck in local optima without making progress. Human demonstrations offer valuable global guidance that helps overcome these challenges, and  $x_{0:T}^{\text{retarget}}$ can naturally serve as the initial guess to CITO-based methods where local adjustments are made to obey dynamical constraints \eqref{eq:pc_dynamics}. \looseness=-1

Thanks to access to the system dynamics $f$ in simulation, we can locally perturb the physical parameters as well as robot and object states around a nominal demonstration. From the single demonstration, we can solve \eqref{eq:predictive_control} for a distribution of tasks with different dynamics $f(x_t, u_t, \theta_t)$, where $\theta_t \sim \rho$ represents all the perturbations. We assume the kinematically retargeted trajectory $x_{0:T}^{\text{retarget}}$ still provides good guidance on achieving the task in the vicinity of the nominal demonstration. This way, a large number of physically consistent trajectories with various physical properties and initial conditions can be generated from a single human demonstration. We outline our data generation pipeline in Algorithm \ref{alg:data_gen}.

\begin{algorithm}
\caption{\textbf{Automated Data Generation}}\label{alg:data_gen}
\textbf{Input:} Probability distribution $\rho$, augmentation number $N$, demo trajectory $x^{\text{demo}}_{0:T}$\;
\textbf{Output:} $N$ dynamically consistent trajectories on target embodiments $\{(x_{0:T}^\star, u_{0:T-1}^\star)\}$\;
${q_{0:T}^{\text{retarget}}}^\star \leftarrow$ Solve \eqref{eq:retargeting_ik} for $x^{\text{demo}}_{0:T}$\;
\For {$n = 1, \ldots, N$} { 
    Sample $\theta_{0:T} \sim \rho$\;
    $(x_{0:T}^\star, u_{0:T-1}^\star) \leftarrow$ Solve \eqref{eq:predictive_control} with $x_{0:T}^{\text{retarget}}, \theta_{0:T}$ and $x_{t+1} = f(x_t, u_t, \theta_t)$\;
}
\end{algorithm}

\begin{table}
        \renewcommand{\arraystretch}{0.8}
        \begin{threeparttable}
        \begin{tabular}{@{}lcc@{}}
        \toprule
        Parameter & Floating Allegro Hand & Bimanual Robot Arms \\
        \midrule
        Init. obj. trans.  pert. (cm) & [$\pm1.5$, $\pm1.5$, 0] & [$\pm 5$, $\pm 5$, 0]\\
        Init. obj. rot. pert. (rad) & [0, 0, $\pm 0.3$] & [0, 0, $\pm 0.3$]\\
        Object side length (cm) & [5.8, 6.2]  & [28, 32] \\
        Object mass (kg) & [0.1, 0.3]  & [0.25, 0.75]  \\
        Friction coefficients & [0.7, 1.3] &  [0.2, 0.4]  \\
        Task horizon (s) & 25 & 50 / 260  (Panda / iiwa) \\
        \bottomrule
        \end{tabular}
        \end{threeparttable}
        \caption{Ranges of different physical parameters $\theta$. The initial object pose is only perturbed in yaw, x, and y to ensure the object sits stably on the table. }
        \label{tab:domain_randomization}
        \vspace{0.5em}
\end{table}

\begin{figure*}[t]
\centering
\includegraphics[width=1.0\textwidth]{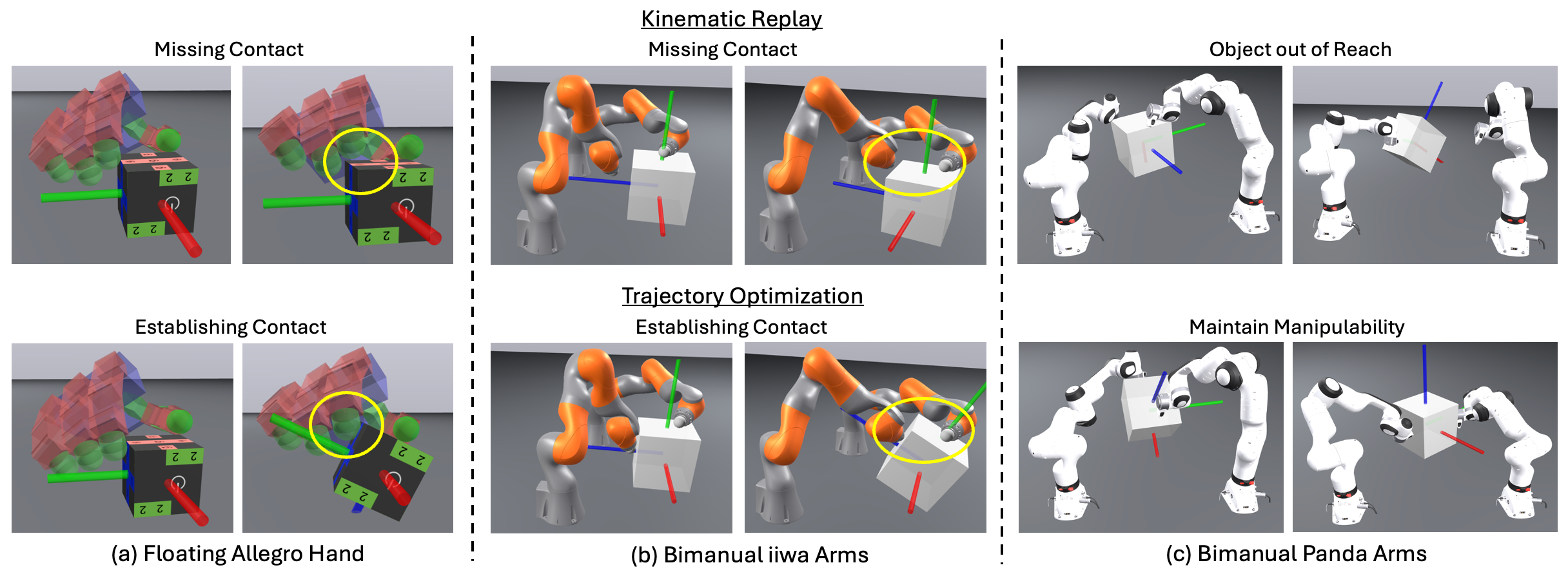}
	\caption{\textbf{Trajectory optimization is crucial for generating dynamically feasible trajectories}. (Top) Before trajectory optimization, the kinematically retargeted demos easily lose contact and drive the object out of reach with different physical parameters or slight deviations in object states. (Bottom) Trajectory optimization encourages robots to establish contact with and maintain good manipulability of the object. The tricolor axis indicates the object orientation.}
	\label{fig:trajopt_unittest}
\end{figure*}

\section{Trajectory Optimization Experiments}

While kinematic retargeting of demonstrations might suffice to generate data for simpler manipulation tasks such as pick and place, it often falls short for the more challenging contact-rich tasks requiring frequent contact mode switches and fine-grained actions. In this section, we demonstrate that trajectory optimization is crucial for generating diverse, dynamically feasible contact-rich trajectories on three high-dimensional dexterous manipulation systems: a floating Allegro hand, bimanual iiwa arms, and bimanual Panda arms.

Our data generation framework is agnostic to the choice of the trajectory optimizer. We implement 
the cross-entropy method (CEM) \cite{de2005tutorial} to solve \eqref{eq:predictive_control} over a distribution of physical parameters and initial conditions, as specified in Table \ref{tab:domain_randomization}. 

\begin{figure*}[t]
    \centering
    \includegraphics[width=1.0\linewidth]{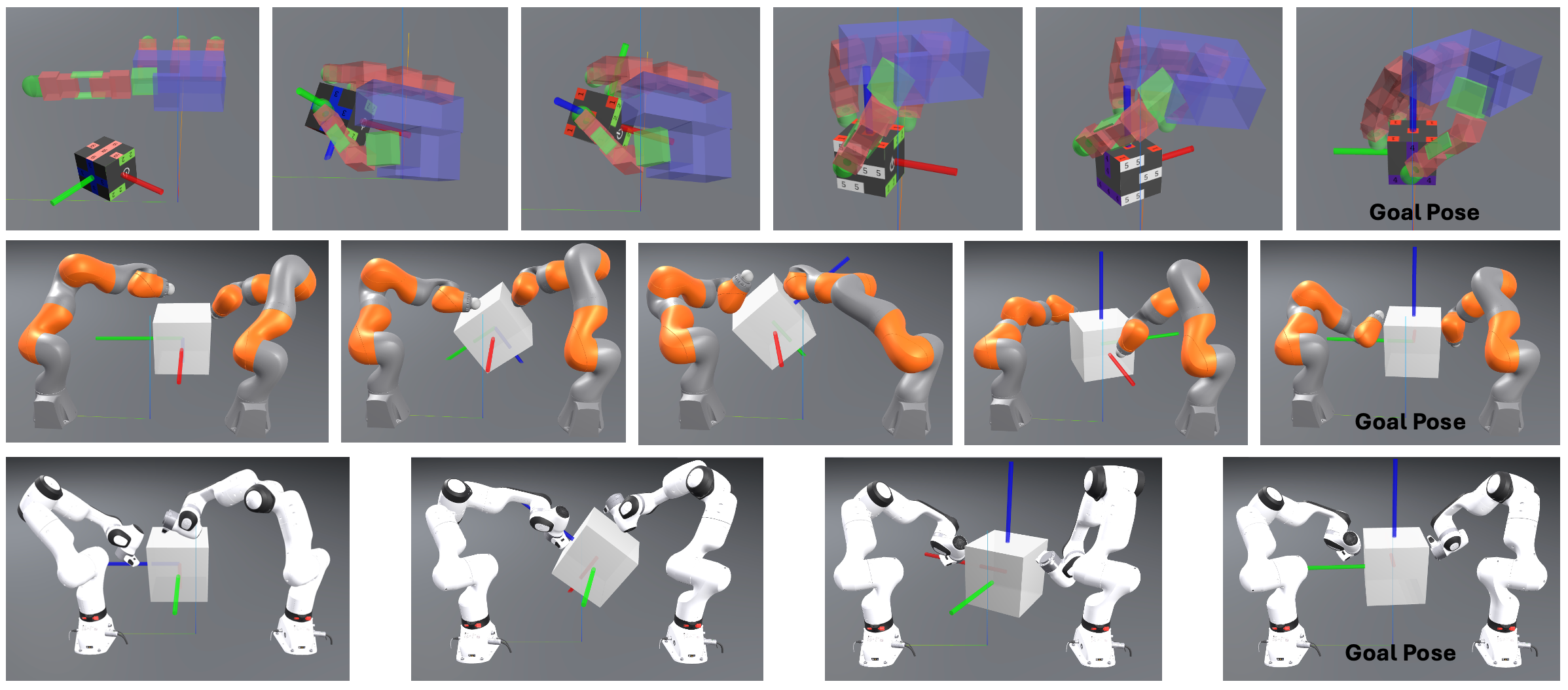}
    \caption{\textbf{Policy rollouts for different embodiments.} The object manipulation task requires the robots to frequently make and break contact with the object. It also requires precise control of the robot since small deviations in positions can result in missing contact interactions and lead to task failure. } 
    \label{fig:policy_rollouts}
\end{figure*}

\underline{\textbf{Task}} Manipulating the object to a target pose on the table (Fig. \ref{fig:policy_rollouts}). The object is initially placed randomly on the table with an arbitrary face upward. Task success is defined as the object reaching within 3 cm and 0.2 rad of the target pose for the Allegro hand, and within 10 cm and 0.2 rad for the bimanual robot arms.  This task requires long-horizon reasoning of complex multi-contact interactions between the robot and the object. The necessary frequent contact mode switches and high-dimensional action space pose great challenges for traditional model-based planners, while the precise contact interactions require fine-grained control actions. 

\begin{table}
\centering
        \renewcommand{\arraystretch}{0.8}
        \begin{threeparttable}
        \begin{tabular}{@{}lcccc@{}}
        \toprule
        Perturbation & Allegro Hand & iiwa Arms & Panda Arms \\
        \midrule
        Original demo &4 / 24 & 5 / 24 & 6 / 24\\
        Object size & 2 / 24 & 1 / 24 & 4 / 24\\
        Initial object translation & 1 / 24 & 3 / 24 & 2 / 24\\
        Initial object orientation & 2 / 24 & 3 / 24& 3 / 24\\
        \midrule
        Trajectory optimization & 2164 / 3000 & 2252 / 3000 & 2462 / 3000 \\
        \bottomrule
        \end{tabular}
        \end{threeparttable}
        \caption{Success rates of replaying kinematically retargeted trajectories of the 24 original human demos, and trajectory optimization under random perturbations in physical parameters and object initial conditions. }
        \label{tab:kin_success_rate}
\end{table}

\begin{figure*}[t]
\centering
\includegraphics[width=1.0\textwidth]{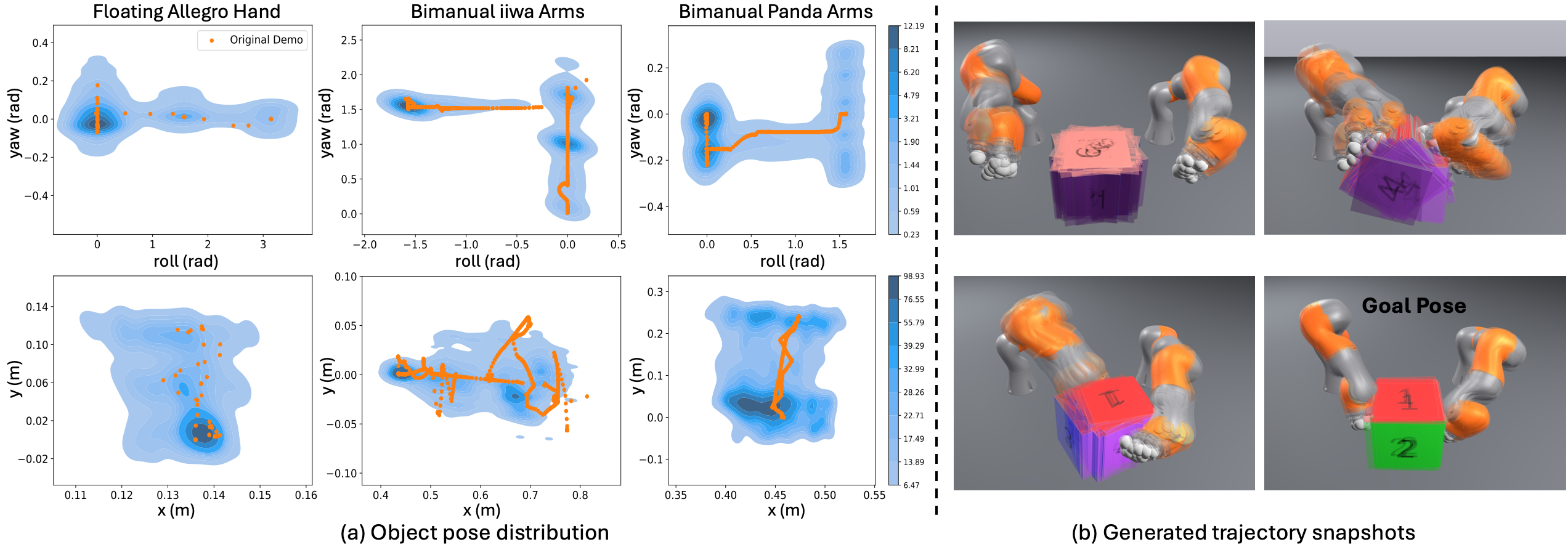}
	\caption{\textbf{Distribution and snapshots of trajectories generated from a single demonstration.} (a) The original demonstration (orange) is locally perturbed and augmented to about 100 dynamically feasible contact-rich trajectories (blue) for each system. The density map represents the object pose distribution of the generated trajectories in the specific 2-dimensional slices. (b) Snapshots of 30 dynamically feasible trajectories under random physical parameters and object initial poses for bimanual iiwa arms are visualized.}
    \label{fig:aug_data_distribution}
\end{figure*}

\underline{\textbf{Dynamic Feasibility}}
While kinematic motion retargeting can generate visually plausible robot and object trajectories, these trajectories often lack dynamical consistency due to the differences in physical parameters and embodiment between the human demonstrator and the target robot. To illustrate this, we replay the kinematically retargeted trajectories of the original 24 human demos and record the success rates for each system in Table \ref{tab:kin_success_rate}. Furthermore, we randomly sample object sizes and perturbations of initial object poses according to Table \ref{tab:domain_randomization} and roll out the nominal kinematically retargeted trajectories. Some trajectories still succeed under certain perturbations thanks to caging grasps or other strategies that encourage robustness during the human demonstration. For all the systems, the successful rollouts are relatively short, manipulating the object to the goal pose within only 1 or 2 rotations. \looseness=-1

The low success rate of purely kinematically retargeted trajectories highlights the importance of trajectory optimization for locally refining the demos for the particular embodiments and physical parameters. Before trajectory optimization, the floating Allegro hand lightly touches the cube and easily loses contact when rotating it clockwise (demonstrated in Fig. \ref{fig:trajopt_unittest}a). After trajectory optimization, the hand increases the contact area, establishing a stable grip for rotation. In Fig. \ref{fig:trajopt_unittest}b, similar behavior that encourages contact can be observed for the bimanual iiwa arms: the demo trajectory tries to rotate the box clockwise only using a single arm, while trajectory optimization encourages the other arm to help hold the box and reorient the box more stably. These refinements that encourage contact are particularly helpful when the object is heavier or smaller, or when the friction coefficients are lower than expected. In addition, replaying the kinematically retargeted trajectory often fails when the object pose deviates slightly from the demonstration, driving the object out of reach (visualized in Fig. \ref{fig:trajopt_unittest}c). In contrast, trajectory optimization 
accounts for the system’s true dynamics and can adjust the robot’s actions accordingly. The success rates of trajectory optimization under random perturbations in physical parameters and object initial conditions for each system are recorded in Table \ref{tab:kin_success_rate}. 

We compare our method to MimicGen \cite{mandlekar2023mimicgen}, an automatic data generation pipeline that adapts demonstrations to diverse spatial configurations through geometric transformations and replay. To ensure a fair comparison, we collect 24 demonstrations directly with the floating Allegro hand in simulation—eliminating embodiment mismatch between the source demonstration and target task—and apply MimicGen-style replay under varying physical parameters (Table I). Pure replay achieves a success rate of $51\%$, while trajectory optimization increases this to $86\%$, and smoothes jitters and imperfections in human demonstrations. Notably, our tasks are highly contact-rich, requiring joint-level control with frequent contact switches. In contrast, the contact behaviors in MimicGen (e.g., pick-and-place, insertion) involve fewer contact switches. Moreover, our task involves \emph{whole-body} manipulation, requiring the entire robot to interact with the object. Simply transforming the end-effector pose in an object-centric manner as in MimicGen disregards the contact between the rest of the robot and the object, and can easily result in loss of contact and task failure.

\underline{\textbf{Cross-Embodiment Generalization}} We demonstrate that a single set of human demonstrations can be effectively repurposed to generate dynamically consistent, contact-rich trajectories across different robotic embodiments with varying task horizons. Specifically, human demonstrations involving two index fingers manipulating a small cube are retargeted to fixed-base bimanual Kuka LBR iiwa and Franka Emika Panda arms manipulating a larger box (visualized in Fig. \ref{fig:kinematic_retargeting}). This approach addresses key challenges in data collection for contact-rich tasks: directly teleoperating two real robot arms to flip a large box would be both physically demanding and cost-prohibitive due to hardware latency, limited feedback, and the embodiment gap--differences in kinematic structure, degrees of freedom, and workspace between human and robotic arms. In contrast, performing the same task on a smaller scale using human fingers is more intuitive, reduces physical effort, and enables faster, more consistent demonstration collection.

The iiwa and Panda arms differ in contact geometry, velocity limits, and joint constraints, all of which are explicitly modeled within the trajectory optimization framework described in \eqref{eq:predictive_control}. For safe hardware deployment, we enforce conservative velocity limits on the iiwa arms, while only applying soft velocity regularization on the Panda arms in simulation to allow for more aggressive motions.

\underline{\textbf{Data Diversity}} 
Trajectory optimization efficiently augments a single demonstration to a wide distribution of trajectories with locally perturbed physical parameters and initial conditions as visualized in Fig. \ref{fig:aug_data_distribution}. The diverse states in the generated dataset cover a larger training distribution and encourage smoother learned policies, as will be discussed in the next section.
\begin{figure*}[t]
    \centering
    \includegraphics[width=1.0\linewidth]{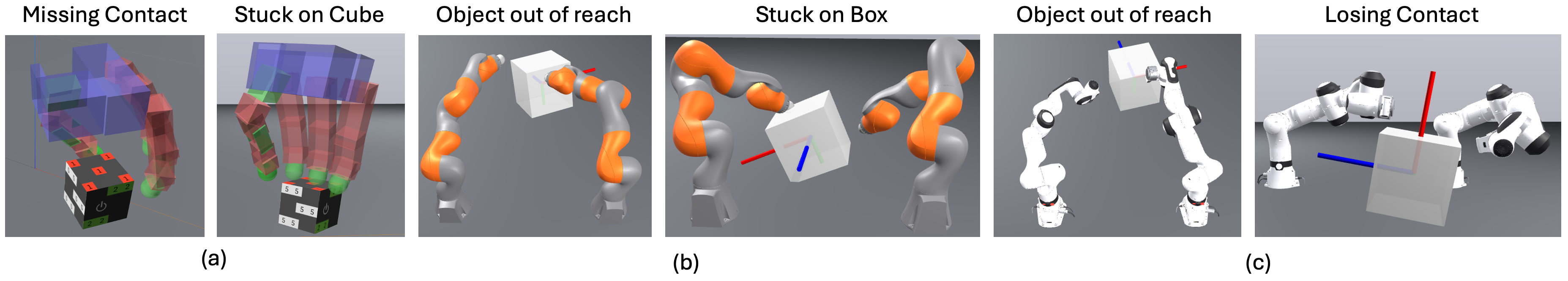}
    \caption{\textbf{Failure cases of baselines.} (a) The baseline policy trained on the original 24 demonstrations for the floating Allegro hand frequently misses contact or gets stuck on the cube. (b-c) The baseline policies for the bimanual robot arms often exhibit jittery motion, resulting in loss of contact, the box being kicked out of reach, or the robot arms running into and getting stuck on the box surface. } 
    \label{fig:policy_failure}
\end{figure*}

\begin{figure}
\centering
\includegraphics[width=0.42\textwidth]{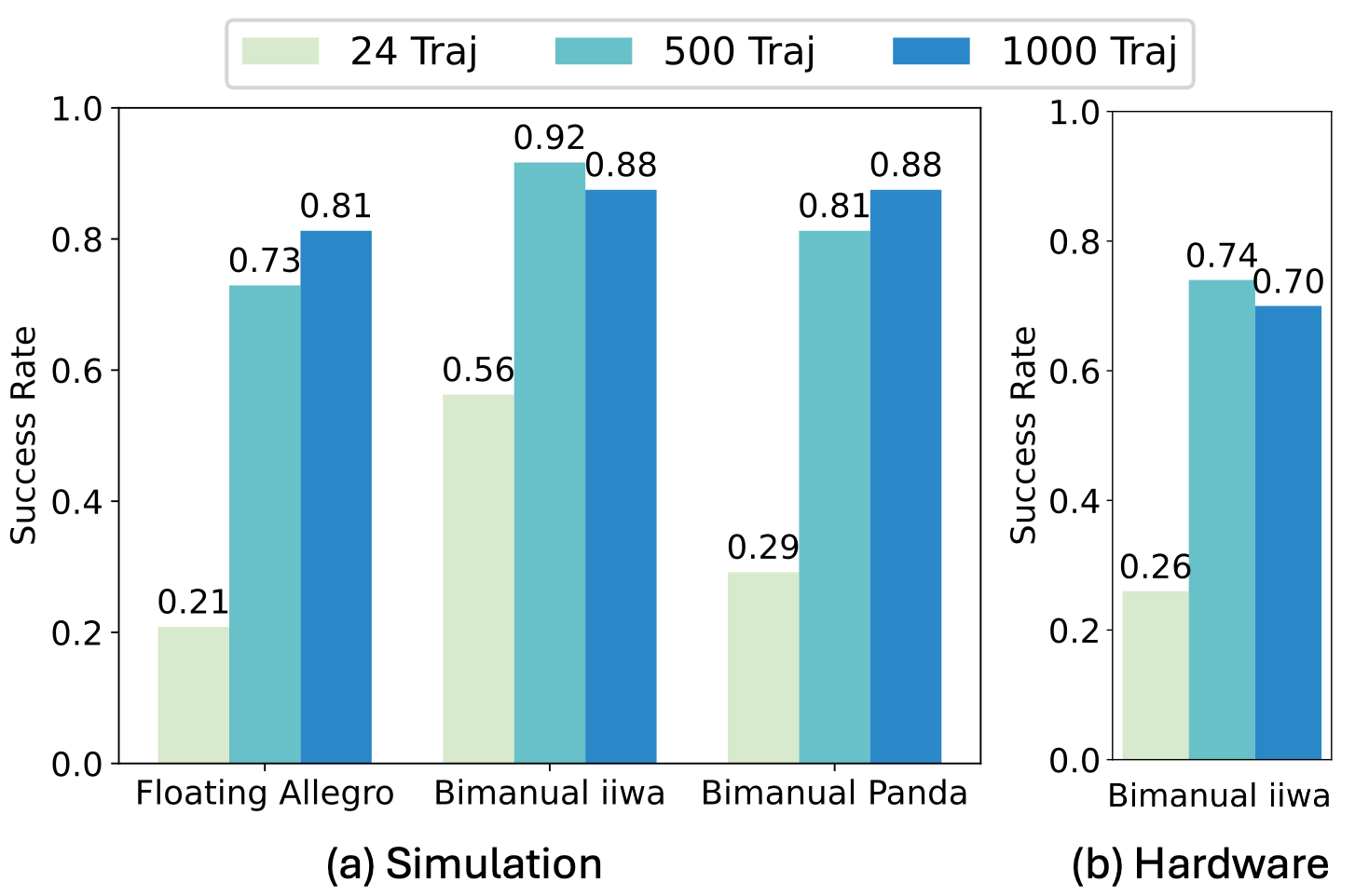}
	\caption{Success rates of policy evaluation in simulation and hardware. }
	\label{fig:success_rate}
    \vspace*{-0.4cm}
\end{figure}


\begin{figure*}[t]
    \centering
    \includegraphics[width=1.0\linewidth]{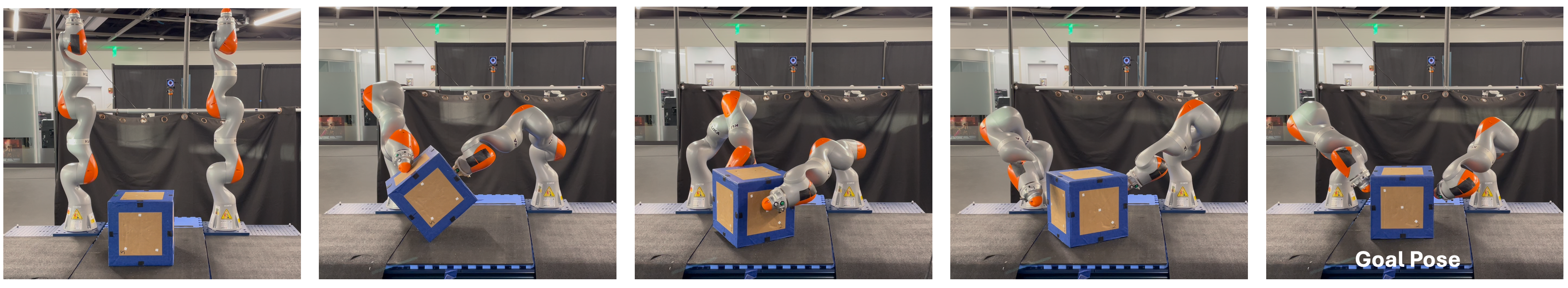}
    \caption{\textbf{Policy rollouts on hardware.} The fixed-base bimanual iiwa arms perform a sequence of coordinated rolling, pitching, and yawing actions to reorient the box to the goal pose. } 
    \label{fig:hardware_rollout}
\end{figure*}

\section{Behavior Cloning Experiments}
We illustrate our framework's capability to efficiently produce diverse, high-quality contact-rich datasets for training behavior cloning policies across multiple robotic platforms, including the floating Allegro hand and the bimanual Panda arms in simulation as well as bimanual iiwa arms on hardware. We show that policies trained on the generated data generalize to a wide distribution of physical parameters and initial conditions, and are much more robust and performant than the ones trained only on the original demonstrations. 
\subsection{Policy Evaluation in Simulation}
\label{subsec:policy_eval_sim}
From only 24 human demonstrations, our data generation pipeline can efficiently generate thousands of dynamically feasible contact-rich trajectories using trajectory optimization. We train state-based diffusion policies \cite{chi2023diffusion} on the 24 original demo trajectories, as well as 500 and 1000 generated trajectories. While our method is compatible with any Behavior Cloning algorithm, we adopt diffusion policies due to its recent success in contact-rich tasks \cite{chi2024universal, zhu2024should, li2024planning}. Fig. \ref{fig:policy_rollouts} visualizes the policy rollouts. We evaluate the performance by conducting 48 policy rollouts for each embodiment in simulation and record the success rates in Fig. \ref{fig:success_rate}. The success criteria are the same as specified in the trajectory optimization experiments.

\subsubsection{Floating Allegro Hand} 
While the human demonstrator completes the task in approximately 5 seconds on average in the virtual reality environment, the demonstration trajectories are temporally scaled by a factor of 2.5 to ensure smoother, dynamically feasible motions on the floating Allegro hand, which is subject to velocity limits. We define the task horizon as 25 seconds to allow the policy sufficient time to recover from missed contacts and other errors during the execution. The task complexity arises from the 22-dimensional action space of the Allegro hand and the long-horizon nature of the task, which requires a sequence of coordinated rolling, pitching, and yawing actions to reorient the cube to an upright position. These factors together present significant challenges for traditional model-based planners without guidance.

The baseline behavior cloning policy trained on the original set of 24 demonstrations achieves a success rate of $10 / 48 = 21\%$ and exhibits significant jittery behavior when encountering out-of-distribution states. The workspace, characterized by diverse object orientations and translations, is sufficiently large that minor deviations during policy rollouts often drive the trajectory out of the demonstrated distribution. Common failure modes include the Allegro hand repeatedly missing contact with the cube or becoming stuck on its surface while attempting reorientation (visualized in Fig. \ref{fig:policy_failure}a), which often result in the object being trapped in intermediate orientations. In contrast, policies trained on the expanded dataset generated by our pipeline demonstrate a higher likelihood of re-establishing contact with the object after initial misses, resulting in significantly improved success rates up to $39 / 48 = 81\%$.

\begin{figure*}[t]
    \centering
    \includegraphics[width=1.0\linewidth]{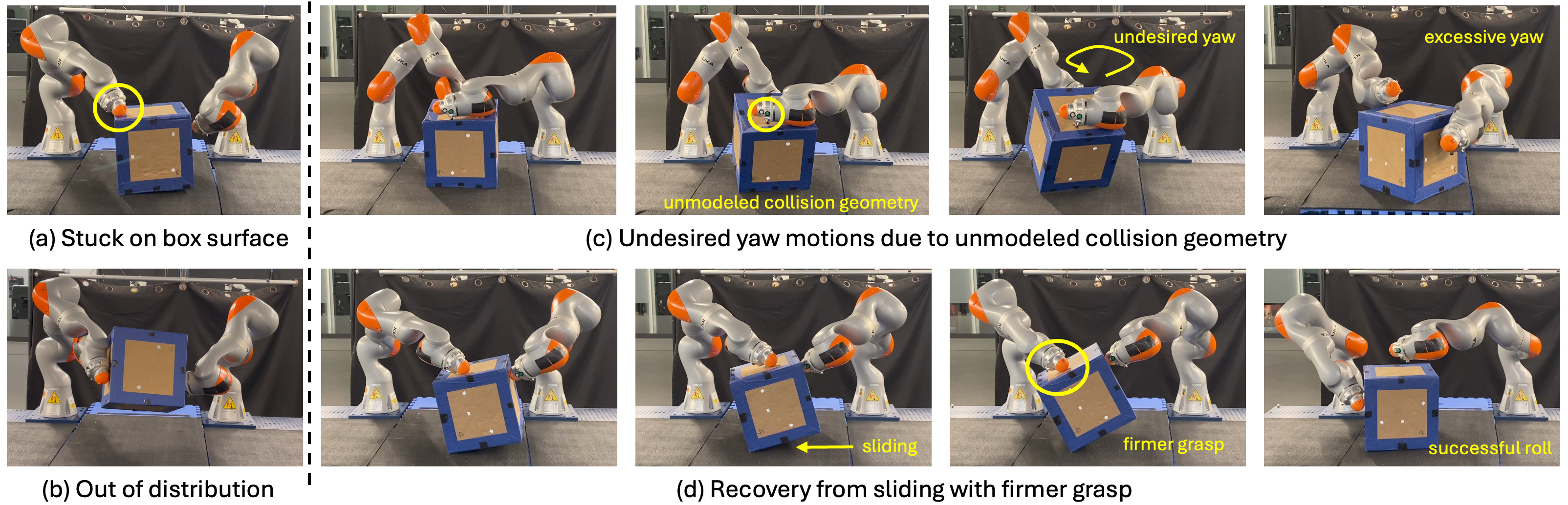}
    \caption{\textbf{Policy failure and recovery on hardware.} The baseline policy frequently (a) gets stuck on the box surface when small deviations from the demonstration trajectories occur, and (b) struggles to recover from out-of-distribution states, where the object is never intentionally lifted for accomplishing the task in the generated dataset. Policies trained on augmented datasets (c) sometimes fail due to unmodeled collision geometry, but (d) can recover from undesired sliding by employing firmer grasps found by trajectory optimization. } 
    \label{fig:hardware_eval}
\end{figure*}
\subsubsection{Bimanual Robot Arms}
The baseline policy trained on the original set of 24 human demonstrations achieves a success rate of $27 / 48 = 56\%$ on the bimanual iiwa system. We hypothesize that the restrictive velocity limits encourage more quasi-static behavior, leading to longer trajectories with a higher density of state-action pairs in the training data. In contrast, the baseline policy yields a success rate of $14/48=29\%$ on the bimanual Panda system, likely due to the more dynamic nature of the learned behavior under its looser velocity constraints. Both baseline policies exhibit remarkably jittery motion, frequently kicking the box out of reach, losing contact, or running into and getting stuck on the box surface during reorientation (visualized in Fig. \ref{fig:policy_failure}b and c). Policies trained on the augmented dataset, however, generate significantly smoother trajectories and are capable of re-establishing contact with the object after initial misses, resulting in as high as $44 / 48 = 92\%$ success rates for bimanual iiwa arms and $42 / 48 = 87.5\%$ for bimanual Panda arms. Additionally, the learned policies capture multimodal behaviors observed in the original human demonstrations, such as rotating the box either clockwise or counterclockwise for similar object poses.

\subsection{Policy Evaluation on Hardware}
We zero-shot deploy the trained policies on hardware for bimanual iiwa arms to flip a 30 cm cubic box on a table (Fig. \ref{fig:hardware_rollout}). An OptiTrack motion capture system is employed to estimate the object pose. The baseline behavior cloning policy only achieves $6/23=26\%$ success rate, with most successful rollouts being relatively short-horizon, involving only 1 or 2 rotations. Common failure modes of the baseline policy include: 1) deviation from the demonstration trajectory, causing the arms to collide with the box surface (Fig. \ref{fig:hardware_eval}a), and 2) significant box sliding during rolling, resulting in the policy encountering out-of-distribution states and failing to recover (Fig. \ref{fig:hardware_eval}b). In contrast, as shown in Fig. \ref{fig:success_rate}b, the policy trained on 500 generated trajectories achieves $17 / 23 = 74\%$ success rate, while the policy trained on 1000 generated trajectories achieves $16/23=70\%$ success rate. Despite occasional box sliding during rolling, these policies demonstrate an improved ability to stabilize the box by using one arm to hold the opposite side more firmly to prevent further sliding (Fig \ref{fig:hardware_eval}d). However, as visualized in Fig \ref{fig:hardware_eval}c, both policies trained on the augmented datasets exhibit failure modes originating from unmodeled collision geometries on iiwa arms, which lead to significant undesired yaw motions of the box during pitch actions.\looseness=-1
\section{Limitations and Future Work}
While our method efficiently generates abundant contact-rich trajectories, several limitations remain. First, although our human-hand demonstration framework is fast and intuitive, it may not fully exploit the kinematic capabilities of the target robot, such as continuous joint rotation or specialized dexterous maneuvers. Future work could explore the application of our automated data generation framework to embodiment-aware legacy datasets, better capturing the unique motion capabilities of different robotic systems.

Second, although our method demonstrates strong performance in the vicinity of the demonstration due to trajectory optimization, the learned policies struggle to recover from states far outside the demonstrated regions, such as those resulting from catastrophic failure. Future work could explore more advanced planning techniques to iteratively improve the learned policies' robustness in unvisited regions of the state space. \looseness=-1

Third, we have demonstrated the effectiveness of our pipeline primarily for training robust state-based policies. Extending the framework to train visuomotor policies by incorporating high-quality synthetic rendering from simulation could further improve policy transferability to real-world scenarios.\looseness=-1

\section{Conclusion}

In this work, we present a novel, cost-effective pipeline that combines physics-based simulations, human demonstrations, and model-based planning to address data scarcity in contact-rich robotic manipulation tasks. A key insight of our approach is that human demonstrations—even when collected on a different morphology—offer valuable global task information that model-based planners often struggle to discover independently due to the high-dimensional search space and complex contact dynamics. By leveraging these demonstrations as a global prior, our method refines and augments them through kinematic retargeting and trajectory optimization, resulting in large datasets of dynamically feasible trajectories across a range of physical parameters, initial conditions, and embodiments. Our framework significantly reduces the reliance on costly, hardware-specific data collection while offering the potential to reuse legacy datasets collected with outdated hardware or configurations. We demonstrate its effectiveness across multiple robotic systems in simulation and successfully zero-shot deploy policies trained on the augmented dataset to a bimanual iiwa hardware setup.
\section{Acknowledgement}
The authors would like to thank Huaijiang Zhu, Haonan Chen and Jiayuan Mao for valuable discussions and insightful feedback on the paper. This work was supported by Robotics and AI Institute Agmd Dtd 8/1/2023, Amazon PO 2D-15694085 and PO 2D-15694048.

\bibliographystyle{IEEEtran}
\bibliography{IEEEabrv,ref}

@inproceedings{chen2024rovi,
  title={RoVi-Aug: Robot and Viewpoint Augmentation for Cross-Embodiment Robot Learning},
  author={Chen, Lawrence Yunliang and Xu, Chenfeng and Dharmarajan, Karthik and Irshad, Zubair and Cheng, Richard and Keutzer, Kurt and Tomizuka, Masayoshi and Vuong, Quan and Goldberg, Ken},
  booktitle={Conference on Robot Learning (CoRL)},
  year={2024}
}

@article{wei2025empirical,
  title={Empirical Analysis of Sim-and-Real Cotraining Of Diffusion Policies For Planar Pushing from Pixels},
  author={Wei, Adam and Agarwal, Abhinav and Chen, Boyuan and Bosworth, Rohan and Pfaff, Nicholas and Tedrake, Russ},
  journal={arXiv preprint arXiv:2503.22634},
  year={2025}
}

@article{chen2023genaug,
  title={Genaug: Retargeting behaviors to unseen situations via generative augmentation},
  author={Chen, Zoey and Kiami, Sho and Gupta, Abhishek and Kumar, Vikash},
  journal={arXiv preprint arXiv:2302.06671},
  year={2023}
}

@article{yu2023scaling,
  title={Scaling robot learning with semantically imagined experience},
  author={Yu, Tianhe and Xiao, Ted and Stone, Austin and Tompson, Jonathan and Brohan, Anthony and Wang, Su and Singh, Jaspiar and Tan, Clayton and Peralta, Jodilyn and Ichter, Brian and others},
  journal={arXiv preprint arXiv:2302.11550},
  year={2023}
}

@inproceedings{zhang2024diffusion,
  title={Diffusion Meets DAgger: Supercharging Eye-in-hand Imitation Learning},
  author={Zhang, Xiaoyu and Chang, Matthew and Kumar, Pranav and Gupta, Saurabh},
  booktitle = {Proceedings of Robotics: Science and Systems},
  year={2024}
}

@InProceedings{tian2024view,
  title =    {View-Invariant Policy Learning via Zero-Shot Novel View Synthesis},
  author =       {Tian, Stephen and Wulfe, Blake and Sargent, Kyle and Liu, Katherine and Zakharov, Sergey and Guizilini, Vitor Campagnolo and Wu, Jiajun},
  booktitle =    {Proceedings of The 8th Conference on Robot Learning},
  pages =    {1173--1193},
  year =     {2025},
  volume =   {270},
  series =   {Proceedings of Machine Learning Research},
  month =    {06--09 Nov},
  publisher =    {PMLR},
}

@misc{drake,
 author = "Russ Tedrake and the Drake Development Team",
 title = "Drake: Model-based design and verification for robotics",
 year = 2019,
 url = "https://drake.mit.edu"
}

@article{li2024drop,
  title={DROP: Dexterous Reorientation via Online Planning},
  author={Li, Albert H and Culbertson, Preston and Kurtz, Vince and Ames, Aaron D},
  journal={arXiv preprint arXiv:2409.14562},
  year={2024}
}

@article{howell2022predictive,
  title={Predictive sampling: Real-time behaviour synthesis with mujoco},
  author={Howell, Taylor and Gileadi, Nimrod and Tunyasuvunakool, Saran and Zakka, Kevin and Erez, Tom and Tassa, Yuval},
  journal={arXiv preprint arXiv:2212.00541},
  year={2022}
}

@article{pezzato2023sampling,
  title={Sampling-based model predictive control leveraging parallelizable physics simulations},
  author={Pezzato, Corrado and Salmi, Chadi and Trevisan, Elia and Spahn, Max and Alonso-Mora, Javier and Corbato, Carlos Hern{\'a}ndez},
  journal={IEEE Robotics and Automation Letters},
  year={2025},
  publisher={IEEE}
}

@article{peng2021amp,
  title={Amp: Adversarial motion priors for stylized physics-based character control},
  author={Peng, Xue Bin and Ma, Ze and Abbeel, Pieter and Levine, Sergey and Kanazawa, Angjoo},
  journal={ACM Transactions on Graphics (ToG)},
  volume={40},
  number={4},
  pages={1--20},
  year={2021},
  publisher={ACM New York, NY, USA}
}

@article{de2005tutorial,
  title={A tutorial on the cross-entropy method},
  author={De Boer, Pieter-Tjerk and Kroese, Dirk P and Mannor, Shie and Rubinstein, Reuven Y},
  journal={Annals of operations research},
  volume={134},
  pages={19--67},
  year={2005},
  publisher={Springer}
}

@article{kurtz2023inverse,
  title={Inverse dynamics trajectory optimization for contact-implicit model predictive control},
  author={Kurtz, Vince and Castro, Alejandro and {\"O}nol, Aykut {\"O}zg{\"u}n and Lin, Hai},
  journal={arXiv preprint arXiv:2309.01813},
  year={2023}
}

@article{posa2014direct,
  title={A direct method for trajectory optimization of rigid bodies through contact},
  author={Posa, Michael and Cantu, Cecilia and Tedrake, Russ},
  journal={The International Journal of Robotics Research},
  volume={33},
  number={1},
  pages={69--81},
  year={2014},
  publisher={Sage Publications Sage UK: London, England}
}

@inproceedings{sleiman2024guided,
  title={Guided Reinforcement Learning for Robust Multi-Contact Loco-Manipulation},
  author={Sleiman, Jean-Pierre and Mittal, Mayank and Hutter, Marco},
  booktitle={8th Annual Conference on Robot Learning (CoRL 2024)},
  year={2024}
}

@article{mordatch2012discovery,
  title={Discovery of complex behaviors through contact-invariant optimization},
  author={Mordatch, Igor and Todorov, Emanuel and Popovi{\'c}, Zoran},
  journal={ACM Transactions on Graphics (ToG)},
  volume={31},
  number={4},
  pages={1--8},
  year={2012},
  publisher={ACM New York, NY, USA}
}

@article{graesdal2024towards,
  title={Towards tight convex relaxations for contact-rich manipulation},
  author={Graesdal, Bernhard Paus and Chia, Shao Yuan Chew and Marcucci, Tobia and Morozov, Savva and Amice, Alexandre and Parrilo, Pablo A and Tedrake, Russ},
  journal={arXiv preprint arXiv:2402.10312},
  year={2024}
}

@article{aydinoglu2024consensus,
  title={Consensus complementarity control for multi-contact mpc},
  author={Aydinoglu, Alp and Wei, Adam and Huang, Wei-Cheng and Posa, Michael},
  journal={IEEE Transactions on Robotics},
  year={2024},
  publisher={IEEE}
}

@article{pomerleau1988alvinn,
  title={Alvinn: An autonomous land vehicle in a neural network},
  author={Pomerleau, Dean A},
  journal={Advances in neural information processing systems},
  volume={1},
  year={1988}
}

@inproceedings{sleiman2019contact,
  title={Contact-implicit trajectory optimization for dynamic object manipulation},
  author={Sleiman, Jean-Pierre and Carius, Jan and Grandia, Ruben and Wermelinger, Martin and Hutter, Marco},
  booktitle={2019 IEEE/RSJ international conference on intelligent robots and systems (IROS)},
  pages={6814--6821},
  year={2019},
  organization={IEEE}
}

@inproceedings{neunert2016efficient,
  title={Efficient whole-body trajectory optimization using contact constraint relaxation},
  author={Neunert, Michael and Farshidian, Farbod and Buchli, Jonas},
  booktitle={2016 IEEE-RAS 16th International Conference on Humanoid Robots (Humanoids)},
  pages={43--48},
  year={2016},
  organization={IEEE}
}

@article{neunert2018whole,
  title={Whole-body nonlinear model predictive control through contacts for quadrupeds},
  author={Neunert, Michael and St{\"a}uble, Markus and Giftthaler, Markus and Bellicoso, Carmine D and Carius, Jan and Gehring, Christian and Hutter, Marco and Buchli, Jonas},
  journal={IEEE Robotics and Automation Letters},
  volume={3},
  number={3},
  pages={1458--1465},
  year={2018},
  publisher={IEEE}
}

@InProceedings{fu2024mobile,
  title =    {Mobile ALOHA: Learning Bimanual Mobile Manipulation using Low-Cost Whole-Body Teleoperation},
  author =       {Fu, Zipeng and Zhao, Tony Z. and Finn, Chelsea},
  booktitle =    {Proceedings of The 8th Conference on Robot Learning},
  pages =    {4066--4083},
  year =     {2025},
  volume =   {270},
  series =   {Proceedings of Machine Learning Research},
  month =    {06--09 Nov},
  publisher =    {PMLR},
}

@article{le2024fast,
  title={Fast contact-implicit model predictive control},
  author={Le Cleac'h, Simon and Howell, Taylor A and Yang, Shuo and Lee, Chi-Yen and Zhang, John and Bishop, Arun and Schwager, Mac and Manchester, Zachary},
  journal={IEEE Transactions on Robotics},
  year={2024},
  publisher={IEEE}
}

@inproceedings{wang2022contact,
  title={Contact-implicit planning and control for non-prehensile manipulation using state-triggered constraints},
  author={Wang, Maozhen and {\"O}nol, Aykut {\"O}zg{\"u}n and Long, Philip and Pad{\i}r, Ta{\c{s}}k{\i}n},
  booktitle={The International Symposium of Robotics Research},
  pages={189--204},
  year={2022},
  organization={Springer}
}

@article{carius2022constrained,
  title={Constrained stochastic optimal control with learned importance sampling: A path integral approach},
  author={Carius, Jan and Ranftl, Ren{\'e} and Farshidian, Farbod and Hutter, Marco},
  journal={The International Journal of Robotics Research},
  volume={41},
  number={2},
  pages={189--209},
  year={2022},
  publisher={SAGE Publications Sage UK: London, England}
}

@inproceedings{tassa2012synthesis,
  title={Synthesis and stabilization of complex behaviors through online trajectory optimization},
  author={Tassa, Yuval and Erez, Tom and Todorov, Emanuel},
  booktitle={2012 IEEE/RSJ International Conference on Intelligent Robots and Systems},
  pages={4906--4913},
  year={2012},
  organization={IEEE}
}

@article{winkler2018gait,
  title={Gait and trajectory optimization for legged systems through phase-based end-effector parameterization},
  author={Winkler, Alexander W and Bellicoso, C Dario and Hutter, Marco and Buchli, Jonas},
  journal={IEEE Robotics and Automation Letters},
  volume={3},
  number={3},
  pages={1560--1567},
  year={2018},
  publisher={IEEE}
}

@inproceedings{moura2022non,
  title={Non-prehensile planar manipulation via trajectory optimization with complementarity constraints},
  author={Moura, Jo{\~a}o and Stouraitis, Theodoros and Vijayakumar, Sethu},
  booktitle={2022 International Conference on Robotics and Automation (ICRA)},
  pages={970--976},
  year={2022},
  organization={IEEE}
}

@inproceedings{onol2019contact,
  title={Contact-implicit trajectory optimization based on a variable smooth contact model and successive convexification},
  author={{\"O}nol, Aykut {\"O}zgun and Long, Philip and Pad{\i}r, Ta{\c{s}}k{\i}n},
  booktitle={2019 International Conference on Robotics and Automation (ICRA)},
  pages={2447--2453},
  year={2019},
  organization={IEEE}
}

@article{hamalainen2015online,
  title={Online control of simulated humanoids using particle belief propagation},
  author={H{\"a}m{\"a}l{\"a}inen, Perttu and Rajam{\"a}ki, Joose and Liu, C Karen},
  journal={ACM Transactions on Graphics (TOG)},
  volume={34},
  number={4},
  pages={1--13},
  year={2015},
  publisher={ACM New York, NY, USA}
}

@article{achiam2023gpt,
  title={Gpt-4 technical report},
  author={Achiam, Josh and Adler, Steven and Agarwal, Sandhini and Ahmad, Lama and Akkaya, Ilge and Aleman, Florencia Leoni and Almeida, Diogo and Altenschmidt, Janko and Altman, Sam and Anadkat, Shyamal and others},
  journal={arXiv preprint arXiv:2303.08774},
  year={2023}
}

@article{khazatsky2024droid,
  title={Droid: A large-scale in-the-wild robot manipulation dataset},
  author={Khazatsky, Alexander and Pertsch, Karl and Nair, Suraj and Balakrishna, Ashwin and Dasari, Sudeep and Karamcheti, Siddharth and Nasiriany, Soroush and Srirama, Mohan Kumar and Chen, Lawrence Yunliang and Ellis, Kirsty and others},
  journal={arXiv preprint arXiv:2403.12945},
  year={2024}
}

@article{touvron2023llama,
  title={Llama: Open and efficient foundation language models},
  author={Touvron, Hugo and Lavril, Thibaut and Izacard, Gautier and Martinet, Xavier and Lachaux, Marie-Anne and Lacroix, Timoth{\'e}e and Rozi{\`e}re, Baptiste and Goyal, Naman and Hambro, Eric and Azhar, Faisal and others},
  journal={arXiv preprint arXiv:2302.13971},
  year={2023}
}

@article{anil2023palm,
  title={Palm 2 technical report},
  author={Anil, Rohan and Dai, Andrew M and Firat, Orhan and Johnson, Melvin and Lepikhin, Dmitry and Passos, Alexandre and Shakeri, Siamak and Taropa, Emanuel and Bailey, Paige and Chen, Zhifeng and others},
  journal={arXiv preprint arXiv:2305.10403},
  year={2023}
}

@InProceedings{kim2024openvla,
  title =    {OpenVLA: An Open-Source Vision-Language-Action Model},
  author =       {Kim, Moo Jin and Pertsch, Karl and Karamcheti, Siddharth and Xiao, Ted and Balakrishna, Ashwin and Nair, Suraj and Rafailov, Rafael and Foster, Ethan P and Sanketi, Pannag R and Vuong, Quan and Kollar, Thomas and Burchfiel, Benjamin and Tedrake, Russ and Sadigh, Dorsa and Levine, Sergey and Liang, Percy and Finn, Chelsea},
  booktitle =    {Proceedings of The 8th Conference on Robot Learning},
  pages =    {2679--2713},
  year =     {2025},
  volume =   {270},
  month =    {06--09 Nov},
  publisher =    {PMLR},
}

@inproceedings{team2024octo,
    title={Octo: An Open-Source Generalist Robot Policy},
    author = {{Octo Model Team} and Dibya Ghosh and Homer Walke and Karl Pertsch and Kevin Black and Oier Mees and Sudeep Dasari and Joey Hejna and Charles Xu and Jianlan Luo and Tobias Kreiman and {You Liang} Tan and Lawrence Yunliang Chen and Pannag Sanketi and Quan Vuong and Ted Xiao and Dorsa Sadigh and Chelsea Finn and Sergey Levine},
    booktitle = {Proceedings of Robotics: Science and Systems},
    address  = {Delft, Netherlands},
    year = {2024},
}

@inproceedings{walke2023bridgedata,
  title={Bridgedata v2: A dataset for robot learning at scale},
  author={Walke, Homer Rich and Black, Kevin and Zhao, Tony Z and Vuong, Quan and Zheng, Chongyi and Hansen-Estruch, Philippe and He, Andre Wang and Myers, Vivek and Kim, Moo Jin and Du, Max and others},
  booktitle={Conference on Robot Learning},
  pages={1723--1736},
  year={2023},
  organization={PMLR}
}

@InProceedings{dasari2019robonet,
  title =    {RoboNet: Large-Scale Multi-Robot Learning},
  author =       {Dasari, Sudeep and Ebert, Frederik and Tian, Stephen and Nair, Suraj and Bucher, Bernadette and Schmeckpeper, Karl and Singh, Siddharth and Levine, Sergey and Finn, Chelsea},
  booktitle =    {Proceedings of the Conference on Robot Learning},
  pages =    {885--897},
  year =   {2020},
  volume =   {100},
  series =   {Proceedings of Machine Learning Research},
  month =    {30 Oct--01 Nov},
  publisher =    {PMLR},
}

@inproceedings{radosavovic2023real,
  title={Real-world robot learning with masked visual pre-training},
  author={Radosavovic, Ilija and Xiao, Tete and James, Stephen and Abbeel, Pieter and Malik, Jitendra and Darrell, Trevor},
  booktitle={Conference on Robot Learning},
  pages={416--426},
  year={2023},
  organization={PMLR}
}

@InProceedings{nair2022r3m,
  title =    {R3M: A Universal Visual Representation for Robot Manipulation},
  author =       {Nair, Suraj and Rajeswaran, Aravind and Kumar, Vikash and Finn, Chelsea and Gupta, Abhinav},
  booktitle =    {Proceedings of The 6th Conference on Robot Learning},
  pages =    {892--909},
  year =   {2023},
  volume =   {205},
  series =   {Proceedings of Machine Learning Research},
  month =    {14--18 Dec},
  publisher =    {PMLR},
}

@article{karamcheti2023language,
  title={Language-driven representation learning for robotics},
  author={Karamcheti, Siddharth and Nair, Suraj and Chen, Annie S and Kollar, Thomas and Finn, Chelsea and Sadigh, Dorsa and Liang, Percy},
  journal={arXiv preprint arXiv:2302.12766},
  year={2023}
}

@inproceedings{damen2018scaling,
  title={Scaling egocentric vision: The epic-kitchens dataset},
  author={Damen, Dima and Doughty, Hazel and Farinella, Giovanni Maria and Fidler, Sanja and Furnari, Antonino and Kazakos, Evangelos and Moltisanti, Davide and Munro, Jonathan and Perrett, Toby and Price, Will and others},
  booktitle={Proceedings of the European conference on computer vision (ECCV)},
  pages={720--736},
  year={2018}
}

@inproceedings{xiang2020sapien,
  title={Sapien: A simulated part-based interactive environment},
  author={Xiang, Fanbo and Qin, Yuzhe and Mo, Kaichun and Xia, Yikuan and Zhu, Hao and Liu, Fangchen and Liu, Minghua and Jiang, Hanxiao and Yuan, Yifu and Wang, He and others},
  booktitle={Proceedings of the IEEE/CVF conference on computer vision and pattern recognition},
  pages={11097--11107},
  year={2020}
}

@inproceedings{chi2024universal,
  title={Universal manipulation interface: In-the-wild robot teaching without in-the-wild robots},
  author={Chi, Cheng and Xu, Zhenjia and Pan, Chuer and Cousineau, Eric and Burchfiel, Benjamin and Feng, Siyuan and Tedrake, Russ and Song, Shuran},
  booktitle = {Proceedings of Robotics: Science and Systems},
  address  = {Delft, Netherlands},
  year = {2024},
}

@article{brockman2016openai,
  title={OpenAI Gym},
  author={Brockman, G},
  journal={arXiv preprint arXiv:1606.01540},
  year={2016}
}

@article{james2020rlbench,
  title={Rlbench: The robot learning benchmark \& learning environment},
  author={James, Stephen and Ma, Zicong and Arrojo, David Rovick and Davison, Andrew J},
  journal={IEEE Robotics and Automation Letters},
  volume={5},
  number={2},
  pages={3019--3026},
  year={2020},
  publisher={IEEE}
}

@inproceedings{wang2024scaling,
 author = {Wang, Lirui and Chen, Xinlei and Zhao, Jialiang and He, Kaiming},
 booktitle = {Advances in Neural Information Processing Systems},
 pages = {124420--124450},
 title = {Scaling Proprioceptive-Visual Learning with Heterogeneous Pre-trained Transformers},
 volume = {37},
 year = {2024}
}

@InProceedings{fu2024humanplus,
  title =    {HumanPlus: Humanoid Shadowing and Imitation from Humans},
  author =       {Fu, Zipeng and Zhao, Qingqing and Wu, Qi and Wetzstein, Gordon and Finn, Chelsea},
  booktitle =    {Proceedings of The 8th Conference on Robot Learning},
  pages =    {2828--2844},
  year =     {2025},
  volume =   {270},
  series =   {Proceedings of Machine Learning Research},
  month =    {06--09 Nov},
  publisher =    {PMLR},
}

@InProceedings{he2024omnih2o,
  title =    {OmniH2O: Universal and Dexterous Human-to-Humanoid Whole-Body Teleoperation and Learning},
  author =       {He, Tairan and Luo, Zhengyi and He, Xialin and Xiao, Wenli and Zhang, Chong and Zhang, Weinan and Kitani, Kris M. and Liu, Changliu and Shi, Guanya},
  booktitle =    {Proceedings of The 8th Conference on Robot Learning},
  pages =    {1516--1540},
  year =     {2025},
  volume =   {270},
  series =   {Proceedings of Machine Learning Research},
  month =    {06--09 Nov},
  publisher =    {PMLR},
}

@inproceedings{mandlekar2023mimicgen,
  title={MimicGen: A Data Generation System for Scalable Robot Learning using Human Demonstrations},
  author={Mandlekar, Ajay and Nasiriany, Soroush and Wen, Bowen and Akinola, Iretiayo and Narang, Yashraj and Fan, Linxi and Zhu, Yuke and Fox, Dieter},
  booktitle={Conference on Robot Learning},
  pages={1820--1864},
  year={2023},
  organization={PMLR}
}

@article{jiang2024dexmimicgen,
  title={DexMimicGen: Automated Data Generation for Bimanual Dexterous Manipulation via Imitation Learning},
  author={Jiang, Zhenyu and Xie, Yuqi and Lin, Kevin and Xu, Zhenjia and Wan, Weikang and Mandlekar, Ajay and Fan, Linxi and Zhu, Yuke},
  journal={arXiv preprint arXiv:2410.24185},
  year={2024}
}

@article{yang2024pushing,
  title={Pushing the limits of cross-embodiment learning for manipulation and navigation},
  author={Yang, Jonathan and Glossop, Catherine and Bhorkar, Arjun and Shah, Dhruv and Vuong, Quan and Finn, Chelsea and Sadigh, Dorsa and Levine, Sergey},
  journal={arXiv preprint arXiv:2402.19432},
  year={2024}
}

@inproceedings{doshiscaling,
  title={Scaling Cross-Embodied Learning: One Policy for Manipulation, Navigation, Locomotion and Aviation},
  author={Doshi, Ria and Walke, Homer Rich and Mees, Oier and Dasari, Sudeep and Levine, Sergey},
  booktitle={8th Annual Conference on Robot Learning}
}

@article{seo2024legato,
  title={LEGATO: Cross-Embodiment Imitation Using a Grasping Tool},
  author={Seo, Mingyo and Park, H Andy and Yuan, Shenli and Zhu, Yuke and Sentis, Luis},
  journal={IEEE Robotics and Automation Letters},
  year={2025},
  publisher={IEEE}
}

@inproceedings{li2024okami,
  title={Okami: Teaching humanoid robots manipulation skills through single video imitation},
  author={Li, Jinhan and Zhu, Yifeng and Xie, Yuqi and Jiang, Zhenyu and Seo, Mingyo and Pavlakos, Georgios and Zhu, Yuke},
  booktitle={8th Annual Conference on Robot Learning},
  year={2024}
}

@inproceedings{wang2023mimicplay,
  title={MimicPlay: Long-Horizon Imitation Learning by Watching Human Play},
  author={Wang, Chen and Fan, Linxi and Sun, Jiankai and Zhang, Ruohan and Fei-Fei, Li and Xu, Danfei and Zhu, Yuke and Anandkumar, Anima},
  booktitle={7th Annual Conference on Robot Learning}
}

@article{chi2023diffusion,
  title={Diffusion policy: Visuomotor policy learning via action diffusion},
  author={Chi, Cheng and Xu, Zhenjia and Feng, Siyuan and Cousineau, Eric and Du, Yilun and Burchfiel, Benjamin and Tedrake, Russ and Song, Shuran},
  journal={The International Journal of Robotics Research},
  pages={02783649241273668},
  year={2023},
  publisher={SAGE Publications Sage UK: London, England}
}

@inproceedings{zhu2023viola,
  title={Viola: Imitation learning for vision-based manipulation with object proposal priors},
  author={Zhu, Yifeng and Joshi, Abhishek and Stone, Peter and Zhu, Yuke},
  booktitle={Conference on Robot Learning},
  pages={1199--1210},
  year={2023},
  organization={PMLR}
}

@InProceedings{zhao2024aloha,
  title =    {ALOHA Unleashed: A Simple Recipe for Robot Dexterity},
  author =       {Zhao, Tony Z. and Tompson, Jonathan and Driess, Danny and Florence, Pete and Ghasemipour, Seyed Kamyar Seyed and Finn, Chelsea and Wahid, Ayzaan},
  booktitle =    {Proceedings of The 8th Conference on Robot Learning},
  pages =    {1910--1924},
  year =     {2025},
  volume =   {270},
  series =   {Proceedings of Machine Learning Research},
  month =    {06--09 Nov},
  publisher =    {PMLR},
}

@ARTICLE{pang2023global,
  author={Pang, Tao and Suh, H. J. Terry and Yang, Lujie and Tedrake, Russ},
  journal={IEEE Transactions on Robotics}, 
  title={Global Planning for Contact-Rich Manipulation via Local Smoothing of Quasi-Dynamic Contact Models}, 
  year={2023},
  volume={39},
  number={6},
  pages={4691-4711},
  doi={10.1109/TRO.2023.3300230}}

@inproceedings{ebert2021bridge,
  title={Bridge data: Boosting generalization of robotic skills with cross-domain datasets},
  author={Ebert, Frederik and Yang, Yanlai and Schmeckpeper, Karl and Bucher, Bernadette and Georgakis, Georgios and Daniilidis, Kostas and Finn, Chelsea and Levine, Sergey},
  booktitle = {Proceedings of Robotics: Science and Systems},
  year={2022}
}

@inproceedings{zhang2018deep,
  title={Deep imitation learning for complex manipulation tasks from virtual reality teleoperation},
  author={Zhang, Tianhao and McCarthy, Zoe and Jow, Owen and Lee, Dennis and Chen, Xi and Goldberg, Ken and Abbeel, Pieter},
  booktitle={2018 IEEE international conference on robotics and automation (ICRA)},
  pages={5628--5635},
  year={2018},
  organization={IEEE}
}

@InProceedings{duan2023ar2,
  title =    {AR2-D2: Training a Robot Without a Robot},
  author =       {Duan, Jiafei and Wang, Yi Ru and Shridhar, Mohit and Fox, Dieter and Krishna, Ranjay},
  booktitle =    {Proceedings of The 7th Conference on Robot Learning},
  pages =    {2838--2848},
  year =     {2023},
  volume =   {229},
  series =   {Proceedings of Machine Learning Research},
  month =    {06--09 Nov},
  publisher =    {PMLR},
}

@INPROCEEDINGS{o2023open,
  author={O'Neill, Abby and Rehman, Abdul and Gupta, Abhinav and Maddukuri, Abhiram and Gupta, Abhishek and Padalkar, Abhishek and Lee, Abraham and Pooley, Acorn and Gupta, Agrim and Mandlekar, Ajay and others},
  booktitle={2024 IEEE International Conference on Robotics and Automation (ICRA)}, 
  title={Open X-Embodiment: Robotic Learning Datasets and RT-X Models : Open X-Embodiment Collaboration0}, 
  year={2024},
  volume={},
  number={},
  pages={6892-6903},
  doi={10.1109/ICRA57147.2024.10611477}}

@article{park2024dexhub,
  title={DexHub and DART: Towards Internet Scale Robot Data Collection},
  author={Park, Younghyo and Bhatia, Jagdeep Singh and Ankile, Lars and Agrawal, Pulkit},
  journal={arXiv preprint arXiv:2411.02214},
  year={2024}
}

@article{chen2024arcap,
  title={Arcap: Collecting high-quality human demonstrations for robot learning with augmented reality feedback},
  author={Chen, Sirui and Wang, Chen and Nguyen, Kaden and Fei-Fei, Li and Liu, C Karen},
  journal={arXiv preprint arXiv:2410.08464},
  year={2024}
}

@article{mandi2022cacti,
  title={Cacti: A framework for scalable multi-task multi-scene visual imitation learning},
  author={Mandi, Zhao and Bharadhwaj, Homanga and Moens, Vincent and Song, Shuran and Rajeswaran, Aravind and Kumar, Vikash},
  journal={arXiv preprint arXiv:2212.05711},
  year={2022}
}

@article{reuters_figure_funding_2024,
  author = {Reuters},
  title = {Robotics startup Figure raises \$67.5 million from Microsoft, Nvidia, and other big tech firms},
  journal = {Reuters},
  year = {2024},
  note = {Accessed: 2024-12-08}
}

@inproceedings{chen2022system,
  title={A system for general in-hand object re-orientation},
  author={Chen, Tao and Xu, Jie and Agrawal, Pulkit},
  booktitle={Conference on Robot Learning},
  pages={297--307},
  year={2022},
  organization={PMLR}
}

@inproceedings{qi2023hand,
  title={In-hand object rotation via rapid motor adaptation},
  author={Qi, Haozhi and Kumar, Ashish and Calandra, Roberto and Ma, Yi and Malik, Jitendra},
  booktitle={Conference on Robot Learning},
  pages={1722--1732},
  year={2023},
  organization={PMLR}
}

@inproceedings{howell2022calipso,
  title={CALIPSO: A differentiable solver for trajectory optimization with conic and complementarity constraints},
  author={Howell, Taylor A and Tracy, Kevin and Le Cleac’h, Simon and Manchester, Zachary},
  booktitle={The International Symposium of Robotics Research},
  pages={504--521},
  year={2022},
  organization={Springer}
}

@article{cheng2023enhancing,
  title={Enhancing dexterity in robotic manipulation via hierarchical contact exploration},
  author={Cheng, Xianyi and Patil, Sarvesh and Temel, Zeynep and Kroemer, Oliver and Mason, Matthew T},
  journal={IEEE Robotics and Automation Letters},
  volume={9},
  number={1},
  pages={390--397},
  year={2023},
  publisher={IEEE}
}

@article{smith2024augmented,
  title={An Augmented Reality Interface for Teleoperating Robot Manipulators: Reducing Demonstrator Task Load through Digital Twin Control},
  author={Smith, Aliyah and Kennedy III, Monroe},
  journal={arXiv preprint arXiv:2409.18394},
  year={2024}
}

@article{nechyporenko2024armada,
  title={ARMADA: Augmented Reality for Robot Manipulation and Robot-Free Data Acquisition},
  author={Nechyporenko, Nataliya and Hoque, Ryan and Webb, Christopher and Sivapurapu, Mouli and Zhang, Jian},
  journal={arXiv preprint arXiv:2412.10631},
  year={2024}
}

@software{vuer,
  author = {Ge Yang},
  title = {{VUER}: A 3D Visualization and Data Collection Environment for Robot Learning},
  version = {},
  publisher = {GitHub},
  url = {https://github.com/vuer-ai/vuer},
  year = {2024}
}

@inproceedings{rajeswaran2017learning,
  title={Learning Complex Dexterous Manipulation with Deep Reinforcement Learning and Demonstrations},
  author={Rajeswaran, Aravind and Kumar, Vikash and Gupta, Abhishek and Vezzani, Giulia and Schulman, John and Todorov, Emanuel and Levine, Sergey},
  booktitle={Robotics: Science and Systems XIV},
  year={2018},
  publisher={Robotics: Science and Systems Foundation}
}

@article{zhu2024should,
  title={Should We Learn Contact-Rich Manipulation Policies from Sampling-Based Planners?},
  author={Zhu, Huaijiang and Zhao, Tong and Ni, Xinpei and Wang, Jiuguang and Fang, Kuan and Righetti, Ludovic and Pang, Tao},
  journal={arXiv preprint arXiv:2412.09743},
  year={2024}
}

@article{li2024planning,
  title={Planning-Guided Diffusion Policy Learning for Generalizable Contact-Rich Bimanual Manipulation},
  author={Li, Xuanlin and Zhao, Tong and Zhu, Xinghao and Wang, Jiuguang and Pang, Tao and Fang, Kuan},
  journal={arXiv preprint arXiv:2412.02676},
  year={2024}
}

@article{vecerik2017leveraging,
  title={Leveraging demonstrations for deep reinforcement learning on robotics problems with sparse rewards},
  author={Vecerik, Mel and Hester, Todd and Scholz, Jonathan and Wang, Fumin and Pietquin, Olivier and Piot, Bilal and Heess, Nicolas and Roth{\"o}rl, Thomas and Lampe, Thomas and Riedmiller, Martin},
  journal={arXiv preprint arXiv:1707.08817},
  year={2017}
}

@article{nasiriany2024robocasa,
  title={RoboCasa: Large-Scale Simulation of Everyday Tasks for Generalist Robots},
  author={Nasiriany, Soroush and Maddukuri, Abhiram and Zhang, Lance and Parikh, Adeet and Lo, Aaron and Joshi, Abhishek and Mandlekar, Ajay and Zhu, Yuke},
  journal={arXiv preprint arXiv:2406.02523},
  year={2024}
}

@inproceedings{nair2018overcoming,
  title={Overcoming exploration in reinforcement learning with demonstrations},
  author={Nair, Ashvin and McGrew, Bob and Andrychowicz, Marcin and Zaremba, Wojciech and Abbeel, Pieter},
  booktitle={2018 IEEE international conference on robotics and automation (ICRA)},
  pages={6292--6299},
  year={2018},
  organization={IEEE}
}

@inproceedings{garrett2024skillmimicgen,
    title={SkillMimicGen: Automated Demonstration Generation for Efficient Skill Learning and Deployment},
    author={Garrett, Caelan and Mandlekar, Ajay and Wen, Bowen and Fox, Dieter},
    booktitle={8th Annual Conference on Robot Learning},
    year={2024}
}

@article{zhu2018reinforcement,
  title={Reinforcement and imitation learning for diverse visuomotor skills},
  author={Zhu, Yuke and Wang, Ziyu and Merel, Josh and Rusu, Andrei and Erez, Tom and Cabi, Serkan and Tunyasuvunakool, Saran and Kram{\'a}r, J{\'a}nos and Hadsell, Raia and de Freitas, Nando and others},
  journal={arXiv preprint arXiv:1802.09564},
  year={2018}
}

@inproceedings{hu2023imitation,
  title={Imitation Bootstrapped Reinforcement Learning},
  author={Hu, Hengyuan and Mirchandani, Suvir and Sadigh, Dorsa},
  booktitle = {Proceedings of Robotics: Science and Systems},
  year={2024}
}

@article{peng2018deepmimic,
  title={Deepmimic: Example-guided deep reinforcement learning of physics-based character skills},
  author={Peng, Xue Bin and Abbeel, Pieter and Levine, Sergey and Van de Panne, Michiel},
  journal={ACM Transactions On Graphics (TOG)},
  volume={37},
  number={4},
  pages={1--14},
  year={2018},
  publisher={ACM New York, NY, USA}
}

@inproceedings{hansen2022modem,
  title={MoDem: Accelerating Visual Model-Based Reinforcement Learning with Demonstrations},
  author={Hansen, Nicklas and Lin, Yixin and Su, Hao and Wang, Xiaolong and Kumar, Vikash and Rajeswaran, Aravind},
  booktitle={The Eleventh International Conference on Learning Representations}
}

@article{belkhale2024data,
  title={Data quality in imitation learning},
  author={Belkhale, Suneel and Cui, Yuchen and Sadigh, Dorsa},
  journal={Advances in Neural Information Processing Systems},
  volume={36},
  year={2024}
}

@misc{ctr,
  title={Dexterous Contact-Rich Manipulation via the Contact Trust Region},
  author={Suh, H.J. Terry and Pang, Tao and Zhao, Tong and Tedrake, Russ},
  year={2025}
}

\clearpage
\section{APPENDIX}
In appendix, we present the implementation details of CEM and policy training.
\subsection{CEM Implementation Details}
\label{sec:appendix_cem}
\begin{table}
\centering
        \renewcommand{\arraystretch}{0.8}
        \begin{threeparttable}
        \begin{tabular}{@{}lccccc@{}}
        \toprule
        Parameter & $T$ & Plan Duration & $q_o$ & $q_r$ & $r_u$ \\
        \midrule
        Floating Allegro Hand & 6 & 1.25 s & 10 & 0.01 & 0.1 \\
        Bimanual iiwa Arms & 6 & 1.25 s & 10 & 0.01 & 10 \\
        Bimanual Panda Arms & 6 & 2.0 s & 10 & 0.01 & 10 \\
        \bottomrule
        \end{tabular}
        \end{threeparttable}
        \caption{\textbf{Parameters for CEM. } $T$: planning horizon. $q_o$: scalar weight for tracking object trajectories. $q_r$: scalar weight for tracking robot trajectories. $r_u$: scalar weight for control input.}
        \label{tab:cem_params}
\end{table}
We provide detailed parameters for the CEM implementation in Table \ref{tab:cem_params}. We optimize over the action knot points $u_{0:T-1}$, which are linearly interpolated to generate action commands sent to Drake. Drake simulates the contact dynamics $f$ at 200 Hz. The state cost matrix  $Q_t = diag(q_o \cdot \mathbf{1}_{n_o}, q_r \cdot \mathbf{1}_{n_r})$, where $n_o$ and $n_r$ denote the object and robot state dimensions, and $\mathbf{1}$ is a vector or all 1's. The terminal state cost matrix $Q_T = 10 \cdot Q_t$. The input cost matrix $R_t = diag(r_u \cdot \mathbf{1}_{n_u})$, where $n_u$ represents the control input dimension. All of the systems adopt 50 samples, 5 elites and initial standard deviation $\sigma = 0.05 \cdot \mathbf{1}_{n_u}$ for action sampling.
\subsection{Policy Implementation Details}
We train UNet-based diffusion policies \cite{chi2023diffusion} for all tasks. The action space is the robot configuration (joint angles, and additional floating base coordinates for the Allegro hand), while the observation space is the robot configuration and object pose (with orientations represented by rotation matrices). Detailed parameters are listed in Table \ref{tab:diffo_po_params}.

\begin{table}
\centering
        \renewcommand{\arraystretch}{0.8}
        \begin{threeparttable}
        \begin{tabular}{@{}lcccccc@{}}
        \toprule
        Parameter & $T_o$ & $T_a$ & Freq & Epochs & Obs. Dim. & Act. Dim. \\
        \midrule
        Floating Allegro Hand & 10 & 40 & 50 & 1000 & 34 & 22 \\
        Bimanual iiwa Arms & 10 & 40 & 20 & 800 & 26 & 14\\
        Bimanual Panda Arms & 10 & 40 & 50 & 800 & 26 & 14\\
        \bottomrule
        \end{tabular}
        \end{threeparttable}
        \caption{\textbf{Parameters for diffusion policies. } $T_o$: observation horizon. $T_a$: action horizon. Freq: environment frequency (Hz, both observations and actions).}
        \label{tab:diffo_po_params}
\end{table}

\end{document}